\documentclass[sigconf]{acmart}
\AtBeginDocument{%
  }

\setcopyright{acmlicensed}
\copyrightyear{2026}
\acmYear{2026}
\acmDOI{XXXXXXX.XXXXXXX}
\acmConference[Conference acronym 'XX]{Make sure to enter the correct
  conference title from your rights confirmation email}{June 03--05,
  2018}{Woodstock, NY}
\acmISBN{978-1-4503-XXXX-X/2018/06}
\renewcommand\footnotetextcopyrightpermission[1]{}
\acmSubmissionID{4701}
\usepackage{amsmath}
\usepackage{booktabs}   
\usepackage{multirow}   

\usepackage{amssymb}    
\usepackage{pifont}     
\usepackage{algorithm}
\usepackage{algpseudocode}
\usepackage[table]{xcolor} 
\newcommand{\greencheck}{\textcolor{green!70!black}{\checkmark}}
\newcommand{\redx}{\textcolor{red!80!black}{\xmark}}
\newcommand{\xmark}{\ding{55}}

\begin{document}

\title{PointVG-R: Internalizing Geometric Reasoning in MLLMs for Precise Pointing Localization via Visual Chain of Thought}
\thanks{\textbf{Code: \url{https://github.com/lingli1724/PointVG-R}}}
\author{Ling Li}
\affiliation{%
  \institution{Tsinghua University}
  \state{Beijing Shi}
  \country{China}}
\email{liling25@mails.tsinghua.edu.cn}

\author{Bowen Liu}
\affiliation{%
  \institution{Dalian University of Technology}
  \state{Dalian Shi}
  \country{China}}
\email{liubw20050702@mail.dlut.edu.cn}

\author{Zinuo Zhan}
\affiliation{%
  \institution{Northwestern Polytechnical University}
  \state{Xi'an Shi}
  \country{China}}
\email{jklwmdwmd@mail.nwpu.edu.cn}

\author{Jianhui Zhong}
\affiliation{%
  \institution{Dalian University of Technology}
  \state{Dalian Shi}
  \country{China}}
\email{zhongjianhui@mail.dlut.edu.cn}

\author{Ziyu Zhu}
\affiliation{%
  \institution{Tsinghua University}
  \state{Beijing Shi}
  \country{China}}
\email{zhuziyu.edward@gmail.com}

\author{Bingcai Wei}
\affiliation{%
  \institution{Wuhan University}
  \state{Wuhan City}
  \country{China}}
\email{weibc97@whu.edu.cn}

\author{Kenglun Chang}
\affiliation{%
  \institution{Dalian University of Technology}
  \state{Dalian Shi}
  \country{China}}
\email{genglun_zhang@apple.com}

\author{Zhidong Deng}
\affiliation{%
  \institution{Dalian University of Technology}
  \state{Dalian Shi}
  \country{China}}
\email{michael@tsinghua.edu.cn}

\renewcommand{\shortauthors}{Trovato et al.}

\begin{abstract}
  Pointing-based visual grounding requires models to precisely locate target objects by deciphering complex spatial relationships between the visual scene and pointing gestures. Traditional methods typically encode input images into static feature representations and perform reasoning primarily within the linguistic domain, often overlooking the rich perceptual cues and explicit spatial geometry inherent in images. In this study, we aim to mitigate the cognitive vulnerability of models in interpreting gestural spatial relations by proposing PointVG-R, a reasoning-guided Multi-modal Large Language Model (MLLM). PointVG-R introduces geometric-aware reasoning for pointing-based grounding, enabling the model to think with images through the strategic integration of Reinforcement Learning (RL) and cold-start data. Specifically, we design a novel geometric reasoning pipeline that simulates the iterative cognitive process humans employ when interpreting pointing gestures. Furthermore, we construct EgoPoint-CoT, a high-quality visual Chain-of-Thought (CoT) dataset featuring detailed reasoning trajectories to guide the model via Supervised Fine-Tuning (SFT) and RL. To address the varying quality of learning signals encountered during training, we further propose an Adaptive Importance Weighting strategy based on Group Variance, which dynamically adjusts reward signals to optimize the learning process. Experimental results demonstrate that PointVG-R achieves SOTA performance, outperforming the baseline by $\textbf{15.86}$ points in mIoU. Extensive ablation studies further validate the efficacy of our proposed modules. Our code and models will be made publicly available.
\end{abstract}

\begin{teaserfigure}
  \includegraphics[width=\textwidth]{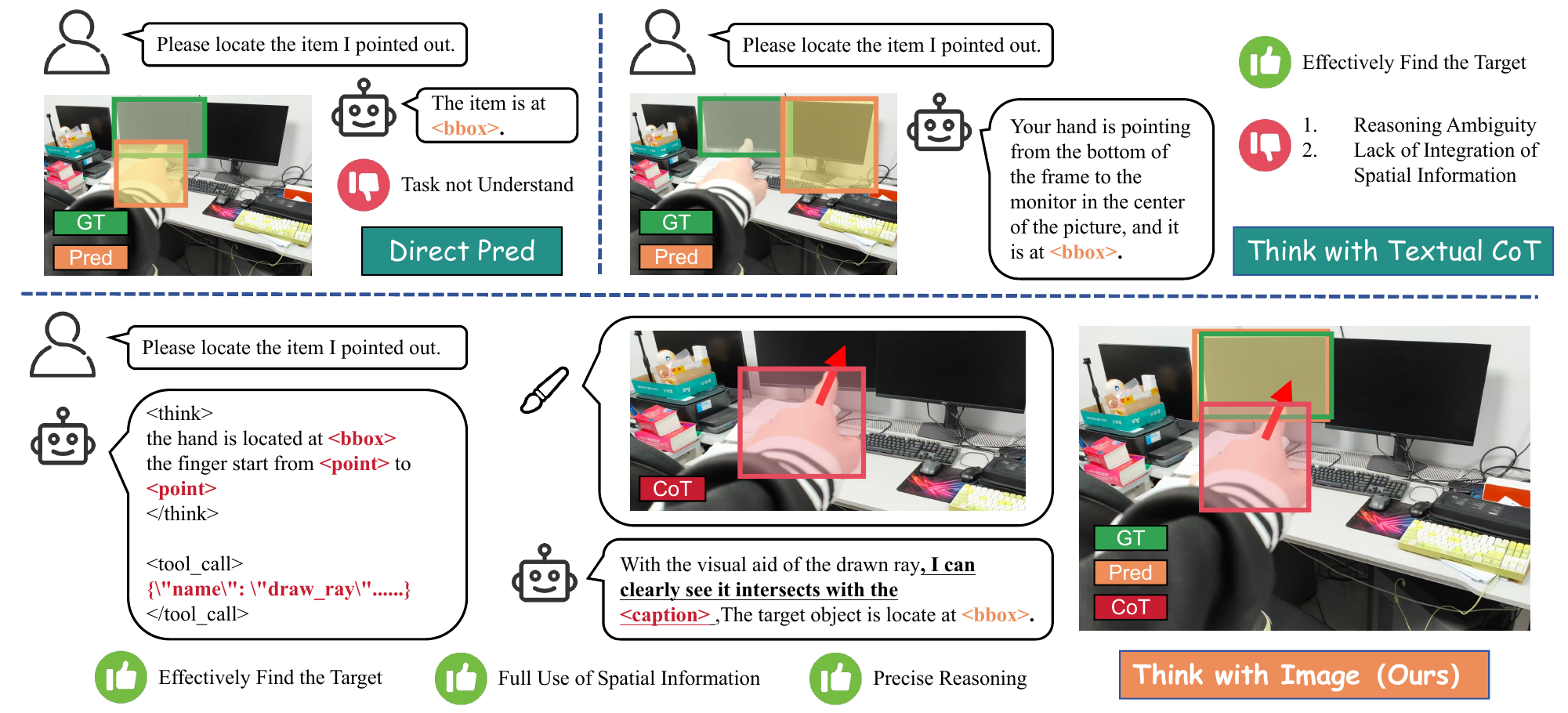}
  \caption{Comparative analysis between PointVG-R and mainstream Multimodal Large Language Models (MLLM) in first-person pointing localization tasks. Above (baseline models): Traditional LVLMs are prone to geometric localization failure, saliency bias, and semantic misinterpretation due to a lack of logical reasoning when processing complex spatial relationships. Below (our method): In stark contrast, PointVG-R achieves a highly interpretable self-explanation process through the Visual Chain-of-Thought (V-CoT). The model not only outputs the final bounding box but also demonstrates a complete logical trajectory from hand localization and geometric ray estimation to target semantic identification. This "thinking with images" paradigm enables it to overcome the black-box limitations of traditional models and achieve more robust spatial understanding.}
  \label{fig:figure1}
\end{teaserfigure}
\received{20 February 2007}
\received[revised]{12 March 2009}
\received[accepted]{5 June 2009}

\maketitle
\section{Introduction}
Visual grounding \cite{zhang2018grounding,dai2024simvg,xiao2024oneref,zhuang2018parallel,xiao2024towardssurvey,anderson2018vision} serves as a fundamental bridge between computer vision and natural language processing \cite{bridge1&geometricreasoning2&rl2_guo2025visual,bridge2_xiao2025towards}, aiming to precisely localize target objects in images based on specific referential expressions \cite{aim_wang2024referencing,zhang2018grounding,xiao2024oneref,liu2024grounding_groundingdino}. While traditional research has primarily focused on semantic descriptions (e.g., "the red cup on the left")\cite{traditional1_wu2023eda,yu2016refcoco,mao2016refcoco+,nagaraja2016refcocog}, gestural pointing provides a more intuitive means of referencing objects within a shared physical environment \cite{bansal2022egoprocess,grauman2022Ego4d,grauman2024Ego-exo4d,chandel2015occlusion,hashi2024systematichand}. This shift toward gesture-based visual grounding \cite{egopoint_li2026languagegroundingreferringexpressions,gesturevg2_chen2021yourefit,gesturevg3_mane2025ges3vig,gesturevg4_guo2025beyond} necessitates that models effectively exploit the physical and geometric cues embedded in visual information to achieve accurate localization \cite{necessitates1_zhu2025struct2d,necessitates2_li2025unfolding}.

We argue that simply treating pointing gestures as another "visual prompt" within existing Multimodal Large Language Model (MLLM) \cite{mllms1_yin2024survey,mllms2_caffagni2024revolution,mllms3_zhang2024mm} architectures is insufficient \cite{insufficient1_nishida2025multimodal,insufficient2_qian2025zoomer}. Standard MLLMs typically encode input images into holistic feature representations and reason primarily through linguistic associations \cite{stdmllm1_liu2025comemo,stdmllm2_qu2026loc3r}. Consequently, they fail to explicitly account for the spatial-geometric constraints inherent in pointing behaviors. 
These models often exhibit "cognitive fragility," leading to hallucinated localization biases in complex interactive scenarios \cite{fail1_shiri2024empirical,fail2_li2025groundingme}.

To address this challenging task, as illustrated in Figure~\ref{fig:figure1}, we propose PointVG-R, a reasoning-guided MLLM that internalizes geometric reasoning \cite{geometricreasoning1_liu2025can,bridge1&geometricreasoning2&rl2_guo2025visual,geometricreasoning3_li202511plus} into the visual grounding process. At the core of our approach is the "Thinking with Images" paradigm \cite{think_with_image1,wei2022chain_cot,thinkingwithimage_wu2025reinforcing}, which advocates for bridging the gap between gesture perception \cite{gestureperception1_jiang2024talon,gestureperception2_deichler2025look} and object localization via a Visual Chain-of-Thought (Visual CoT) \cite{vcot1_shao2024visual,vcot2_man2025argus,vcot3_zhao2025cot,vcot4_ni2024visual}. Rather than performing one-step direct predictions, PointVG-R is trained to execute a human-like, iterative cognitive process: identifying pointing cues, projecting implicit geometric trajectories, and anchoring the target through spatial reasoning. To facilitate this, we constructed a pioneering dataset, EgoPoint-CoT \cite{egopoint_li2026languagegroundingreferringexpressions}, featuring fine-grained spatial reasoning trajectories. We employ a two-stage training protocol combining Supervised Fine-Tuning (SFT) \cite{sft1_bai2025univg,sft2_he2024improved,sft3_lai2025survey} and Reinforcement Learning (RL) \cite{rl1_zhou2025learning,bridge1&geometricreasoning2&rl2_guo2025visual,rl3,rl4_mienye2026deep} to ensure the model's internal reasoning aligns with objective spatial logic.

During the reinforcement learning phase, we define a reward function with multi-dimensional verifiable rewards to provide precise geometric feedback tailored to gesture-based grounding. In this process, we identify a critical bottleneck in applying RL to localization: the heterogeneity in learning signal quality \cite{heterogeneity1_pang2026ica,heterogeneity2_jiao2026credit}. Due to the stochastic nature of sampling in RL, different sample groups offer varying levels of optimization value, where noisy samples can severely destabilize the training of precise bounding box (BBox) coordinates \cite{noise1_waite2025rls3,noise2_cui2024exploring}. To resolve this, we introduce a Group-Variance-based Adaptive Importance Weighting strategy. By dynamically recalibrating the contribution of reward signals based on intra-group consistency, we significantly enhance both training stability and localization precision.

Experimental evaluations on egocentric benchmarks demonstrate that PointVG-R establishes a new state-of-the-art (SOTA) performance, surpassing competing baselines by $15.86\%$. Our results underscore the necessity of explicit geometric reasoning for robust gesture-based grounding. Our contributions include:

(1) \textbf{PointVG-R:} We propose PointVG-R, a reasoning-guided MLLM that enhances gesture-based grounding by internalizing explicit spatio-geometric reasoning.
(2) \textbf{Paradigm and Dataset:} We introduce the "Thinking with Images" paradigm and the EgoPoint-CoT dataset, utilizing exhaustive visual reasoning chains to guide the model’s iterative cognitive process.
(3) \textbf{Algorithmic Optimization:} We design multi-dimensional geometric rewards and a group-variance-based adaptive weighting strategy to address the challenges of signal quality variance and training instability in RL-based coordinate regression.
(4) \textbf{SOTA Performance:} PointVG-R achieves superior results across relevant tasks. We will release our code, dataset, and model weights to the public.

\begin{figure*}
    \centering
    \includegraphics[width=\textwidth]{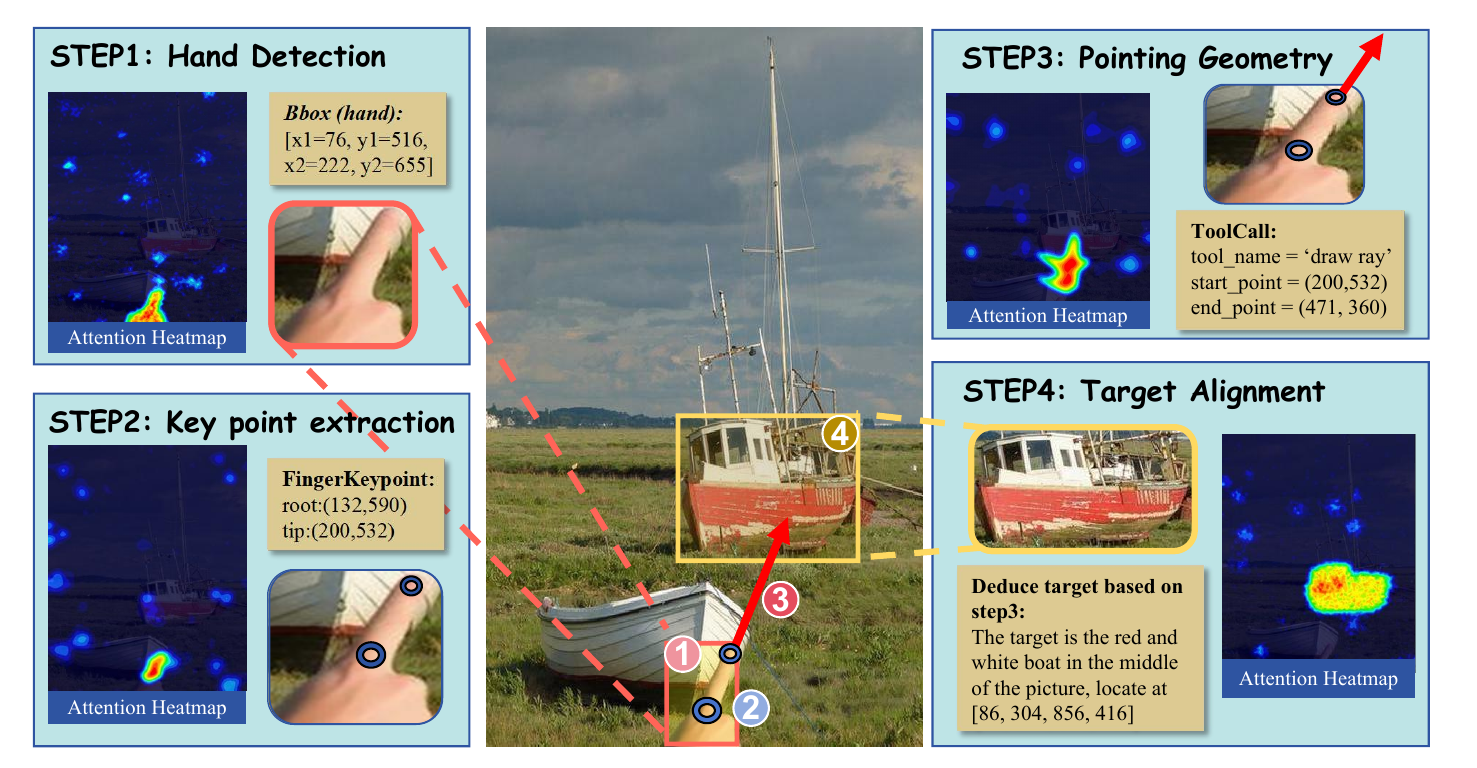}
    \caption{Visual Chain-of-Thought (V-CoT) reasoning trajectory of PointVG-R for pointing localization. The process comprises four sequential steps: (1) hand detection, (2) fingertip keypoint extraction, (3) pointing geometry (ray drawing and endpoint prediction), and (4) target bounding box alignment. Each step is accompanied by its corresponding attention heatmap, visualizing where the model focuses during inference.}
    \label{fig:figure2}
\end{figure*}

\section{Method}
\subsection{Problem Formulation}

\noindent \textbf{Task Definition and Notation.} 
Given an egocentric image $I \in \mathbb{R}^{H \times W \times 3}$ containing a human pointing gesture $G$, the objective of this study is to learn a parameterized model $f_\theta$ that accurately localizes the target object referenced by the gesture. We define the output as a spatial bounding box $b = [x_1, y_1, x_2, y_2] \in [0, 1]^4$, where $(x_1, y_1)$ and $(x_2, y_2)$ denote the normalized coordinates of the top-left and bottom-right corners, respectively.

\vspace{5pt}
\noindent \textbf{Probabilistic Reasoning Modeling.} 
Differing from conventional end-to-end direct mapping, we formulate the interpretation of pointing behaviors as an explicit spatial reasoning process. To this end, we introduce a set of intermediate reasoning variables $R = \{r_1, r_2, \dots, r_n\}$, referred to as the \textit{Visual Chain-of-Thought (CoT)}. The grounding task is thus modeled as a joint probability distribution over the target box $b$ and the reasoning trajectory $R$:
\begin{equation}
P(b, R | I; \theta) = P(R | I; \theta) \cdot P(b | I, R; \theta)
\end{equation}
In this framework, the model first generates a logically consistent reasoning trajectory $R$ based on visual cues in $I$, and subsequently determines the geometric boundary $b$ guided by $R$. This explicit modeling replaces black-box mapping with structured spatial parsing, enhancing robustness against ambiguity.

\vspace{5pt}
\noindent \textbf{Cold-start Knowledge Injection (SFT).} 
To provide the model with a stable initial policy and mitigate the sparsity of rewards in reinforcement learning, we first perform Supervised Fine-Tuning (SFT) as a cold-start phase. Utilizing a dataset $\mathcal{D}$ with expert-level reasoning annotations $(R^*, b^*)$, we minimize the negative log-likelihood loss to inject foundational geometric priors into $\theta$:
\begin{equation}
\mathcal{L}_{SFT}(\theta) = -\mathbb{E}_{(I, R^*, b^*) \sim \mathcal{D}} \left[ \log P(R^*, b^* | I; \theta) \right]
\end{equation}

\vspace{5pt}
\noindent \textbf{Alignment and Enhancement Phase (RL).} 
Upon establishing a reliable baseline via cold-start, we further optimize the model via Policy Gradient \cite{PolicyGradient1_sutton1999policy,PolicyGradient2_lehmann2024definitive} to align its reasoning with diverse spatial scenarios. To address the non-uniform signal quality in sampled trajectories, we introduce an \textbf{Adaptive Importance Weight} $w_g$. The optimization objective $J(\theta)$ is defined as:
\begin{equation}
J(\theta) = \mathbb{E}_{I \sim \mathcal{D}, b \sim P_\theta} \left[ w_g \cdot \mathcal{R}(b, b^*) \cdot \nabla_\theta \log P(b, R | I; \theta) \right]
\end{equation}
where $\mathcal{R}$ is the reward function integrating localization accuracy and reasoning consistency. The weight $w_g$ is dynamically computed based on \textit{Group Variance}, effectively attenuating the interference of low-value or noisy samples on gradient updates, ensuring that $\theta$ converges toward the optimal reasoning and grounding path.
\begin{figure*}
    \centering
    \includegraphics[width=\textwidth]{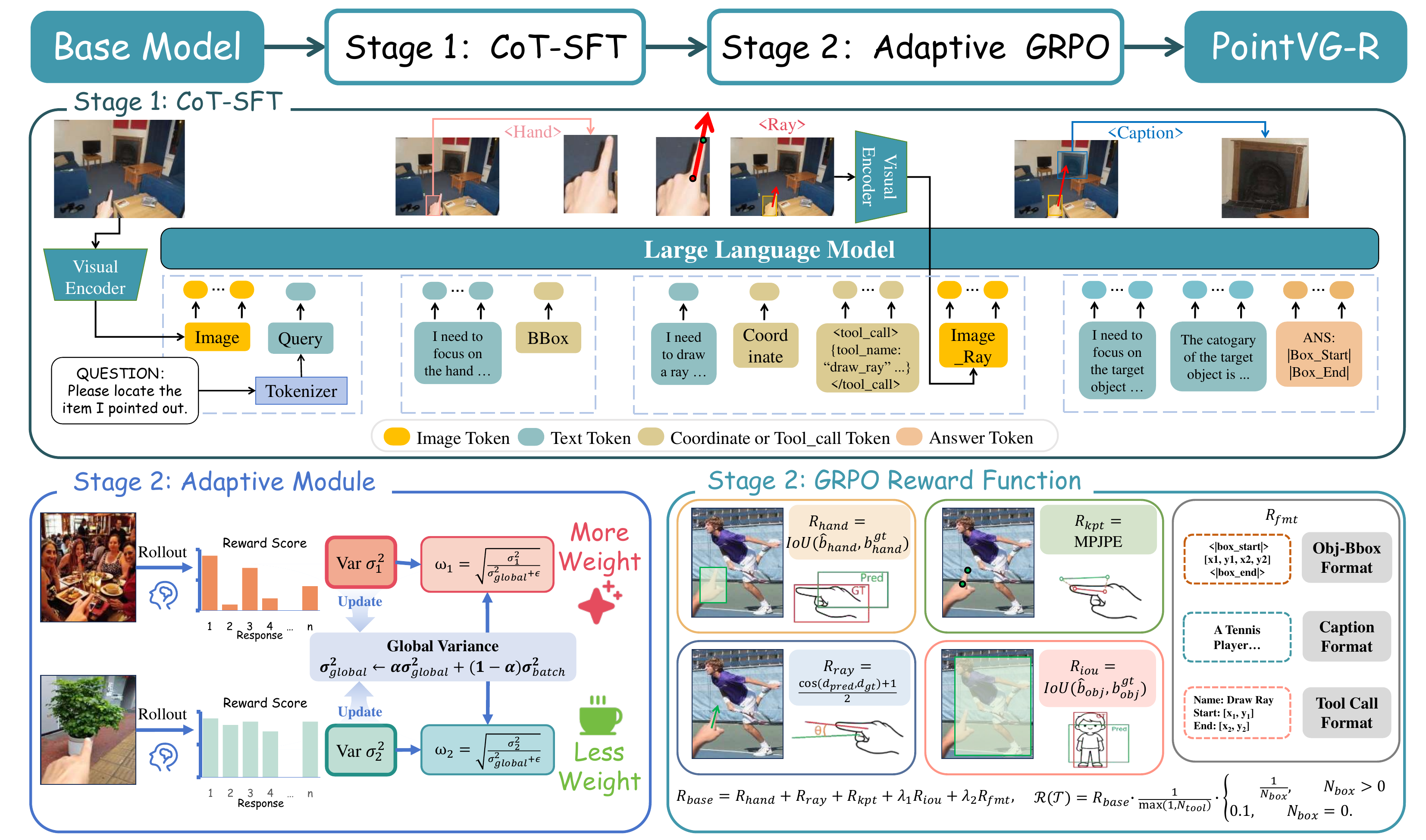}
    \caption{Overview of the PointVG-R system architecture. (a) The first stage injects human-like iterative reasoning priors through CoT-SFT; (b) the second stage adopts an improved GRPO algorithm to achieve efficient optimization through group-variance adaptive weights; (c) structured reward functions designed for the intermediate trajectories of V-CoT. This architecture achieves a leap from primitive spatial perception to high-level logical alignment through two-stage synergy.}
    \label{fig:figure3}
\end{figure*}

\subsection{Model Architecture}


The architecture of PointVG-R is built upon the \textbf{Qwen2.5-VL-Instruct} \cite{bai2025qwen25vltechnicalreport} multimodal foundation model, consisting of a visual encoder $\mathcal{E}_V$ and an LLM backbone $\mathcal{E}_L$. 
To enable explicit spatial geometry parsing, the model processes inputs through a four-stage Visual Chain-of-Thought, coupled with an image-ray re-encoding module. 
As illustrated in Figure~\ref{fig:figure3}, PointVG-R is trained using a two-stage framework that includes CoT-SFT warm-up followed by Adaptive GRPO alignment.

\vspace{5pt}
\noindent \textbf{2.2.1 Structured V-CoT and Feature Re-encoding.} \\
Given an image $I$ and query $Q$, we first extract global visual features $F_V = \mathcal{E}_V(I)$. To explicitly model the visual chain-of-thought (V-CoT), we augment the tokenizer with structured spatial anchors, enabling step-wise geometric reasoning via the decoder $\mathcal{E}_L$. As illustrated in Fig.~\ref{fig:figure2}, the reasoning process is decomposed into four sequential stages.

\vspace{5pt}
\noindent \textbf{Step 1: Hand Detection.} 
We localize the interaction source by predicting the hand bounding box $r_1 = b_{\text{hand}}$, enclosed by \texttt{<|box\_start|>} and \texttt{<|box\_end|>} tokens.

\vspace{5pt}
\noindent \textbf{Step 2: Keypoint Extraction.} 
The model predicts finger root and tip keypoints $r_2 = (\mathbf{p}_{\text{root}}, \mathbf{p}_{\text{tip}})$ to parameterize the pointing direction, structured by \texttt{<|point\_start|>} and \texttt{<|point\_end|>}.

\vspace{5pt}
\noindent \textbf{Step 3: Pointing Geometry.} 
A ray is constructed from $r_2$ to guide spatial sampling. We extract a ray-aligned region $I_{\text{ray}} = \Phi(I, r_2)$ and fuse its features with global context:
\[
F_V' = F_V \oplus \mathcal{E}_V(I_{\text{ray}}).
\]

\vspace{5pt}
\noindent \textbf{Step 4: Target Alignment.} 
Conditioned on geometry and enriched features $F_V'$, the model predicts the final target box $r_4 = b$, again using \texttt{<|box\_start|>} and \texttt{<|box\_end|>} tokens. Structural transitions are regulated by \texttt{<tool\_call>} tokens.

\vspace{5pt}
\noindent The overall process is factorized as:
\vspace{-4pt}
\begin{equation}
\begin{split}
P(b, R \mid I, Q; \theta) &= P(r_1|F_V) P(r_2|F_V, r_1) \\
&\quad \cdot P(r_3|F_V, r_{1:2}) P(r_4|F_V', r_{1:3}),
\end{split}
\end{equation}
where $R=\{r_1,r_2,r_3\}$ and $F_V, F_V'$ denote global and geometry-enhanced features.

\noindent \textbf{2.2.2 Two-stage Adaptive Joint Framework.}
 \label{subsubsec:Two_stageAdaptiveJointFramework} \\
To drive the parameters $\theta$ to stably converge toward the optimal geometric deduction path, the architectural evolution is divided into two stages. Throughout this process, to prevent catastrophic forgetting and reduce memory overhead, we \textbf{keep the parameters of the visual encoder $\mathcal{E}_V$ completely frozen} to preserve its pre-trained generalized perception. Gradients are only backpropagated through the vision-language projector and $\mathcal{E}_L$.

\noindent \textbf{Cold-start Knowledge Injection (SFT).} Rather than relying on free-form text generation, which is prone to parsing errors during RL, the rigorous formatting protocol described above transforms the raw reasoning text into a highly deterministic decision sequence. During the cold-start phase, we apply Low-Rank Adaptation (LoRA) \cite{lora1_hu2022lora,lora2_yang2024low} for parameter-efficient fine-tuning of $\mathcal{E}_L$. The model absorbs this rigid syntax and spatial reasoning pattern on the EgoPoint-CoT dataset, providing an optimal initial policy distribution $\pi_{\text{ref}}$ and tightly constraining the generation space.

\noindent \textbf{Adaptive RL with GRPO \cite{grpo1_shao2024deepseekmath,grpo2_zhou2026demystifying}.} In the policy alignment phase, the rigid syntax established in Stage 1 enables the precise, rule-based calculation of the format reward $R_{fmt}$. To address reward noise caused by coordinate fluctuations during RL sampling, we embed an Adaptive Importance Weight module into the gradient backpropagation path. This module computes the reward variance within the sampled group in real-time to dynamically generate a weight $w_g \propto \sqrt{\text{Var}(\mathcal{R}_G)}$. By penalizing the gradient contribution of high-variance (i.e., unstable and noisy) groups, the joint training mechanism ensures that the model robustly filters invalid explorations and converges to a physically consistent optimal solution.
\begin{figure*}
    \centering
    \includegraphics[width=\textwidth]{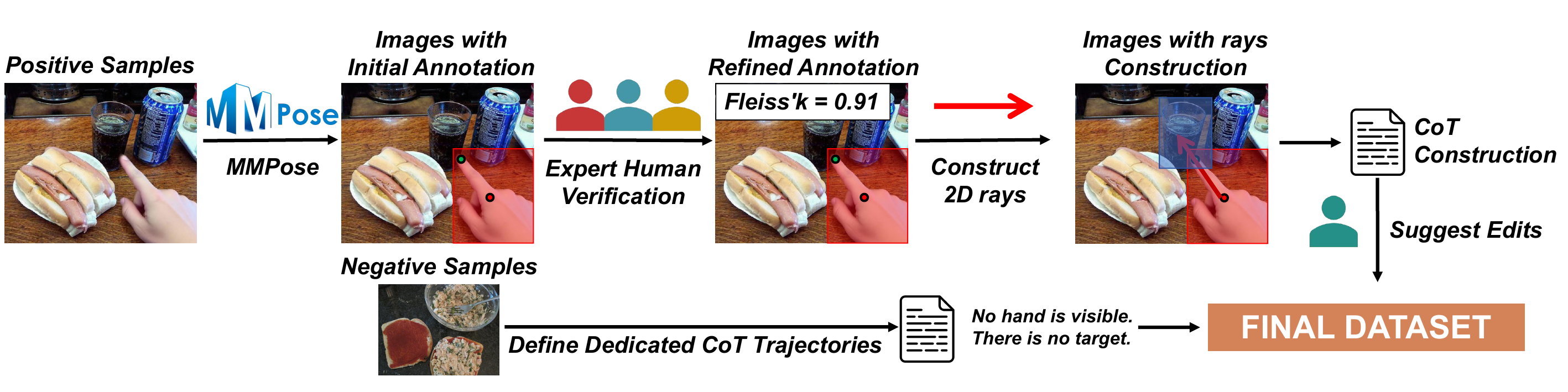}
    \caption{Construction pipeline of the EgoPoint-CoT training dataset.}
    \label{fig:figure4}
\end{figure*}
\subsection{Reward Design}
We design a structured reward to evaluate the generated V-CoT trajectory from both geometric accuracy and output validity. 
The reward is defined hierarchically as
\[
R_{\text{base}} = \lambda_2 R_{\text{acc}} + R_{\text{fmt}},
\]
where \(R_{\text{acc}}\) measures geometric prediction accuracy, and \(R_{\text{fmt}}\) encourages a valid output structure.

\paragraph{Accuracy Reward.}
The accuracy reward is further defined as
\[
R_{\text{acc}} = \lambda_1 R_{\text{iou}} + R_{\text{toolcall}},
\]
where \(R_{\text{iou}}\) evaluates the final target grounding, and \(R_{\text{toolcall}}\) measures the quality of the intermediate pointing process.

The target box reward is computed as the IoU between the final predicted object box and the ground-truth target box, where the final predicted object box is defined as the last box appearing after the final tool call:
\[
R_{\text{iou}} = \mathrm{IoU}(\hat{b}_{\text{obj}}, b^{\text{gt}}_{\text{obj}}).
\]

The tool-call reward is composed of keypoint, hand-box, and ray rewards:
\[
R_{\text{toolcall}} = R_{\text{kpt}} + R_{\text{hand}} + R_{\text{ray}}.
\]

Specifically, \(R_{\text{hand}}\) evaluates the first predicted box in the trajectory, treated as the hand box, using IoU with the ground-truth hand box. 
\(R_{\text{ray}}\) measures the angular consistency between the predicted ray from the \texttt{draw\_ray} tool call and the ground-truth ray, and normalizes the reward to the range \([0,1]\). 
\(R_{\text{kpt}}\) supervises two pointing keypoints using scale-normalized MPJPE \cite{mpjpe1_toshpulatov2022human}:
\[
R_{\text{kpt}} =
\mathrm{clip}\!\left(
1-\frac{1}{K D_{\text{hand}}}
\sum_{k=1}^{K}\|\hat{p}_k-p_k^{\text{gt}}\|_2,0,1
\right), \quad K=2.
\]

\paragraph{Format Reward.}
The format reward \(R_{\text{fmt}}\) checks whether the response contains a valid final grounding structure, including both a predicted object box after the final tool call and a valid caption reference.

\paragraph{Repetition Penalty.}
To discourage redundant reasoning steps, we penalize repeated ray tool calls and repeated object box predictions. 
Let \(N_{\text{tool}}\) denote the number of draw\_ray calls and \(N_{\text{box}}\) the number of object boxes generated after the final tool call. 
The final reward is given by
\[
\mathcal{R}(\mathcal{T}) =
R_{\text{base}}
\cdot \frac{1}{\max(1,N_{\text{tool}})}
\cdot
\begin{cases}
\frac{1}{N_{\text{box}}}, & N_{\text{box}}>0 \\
0.1, & N_{\text{box}}=0 .
\end{cases}
\]

\paragraph{Negative Samples.}
For images without valid pointing annotations, we apply a penalty-only rule. 
If the model still produces tool calls, boxes, point tokens, or caption spans, corresponding penalties are applied to suppress hallucinated reasoning traces.

\begin{figure*}[!h]
    \centering
    \includegraphics[width=\textwidth]{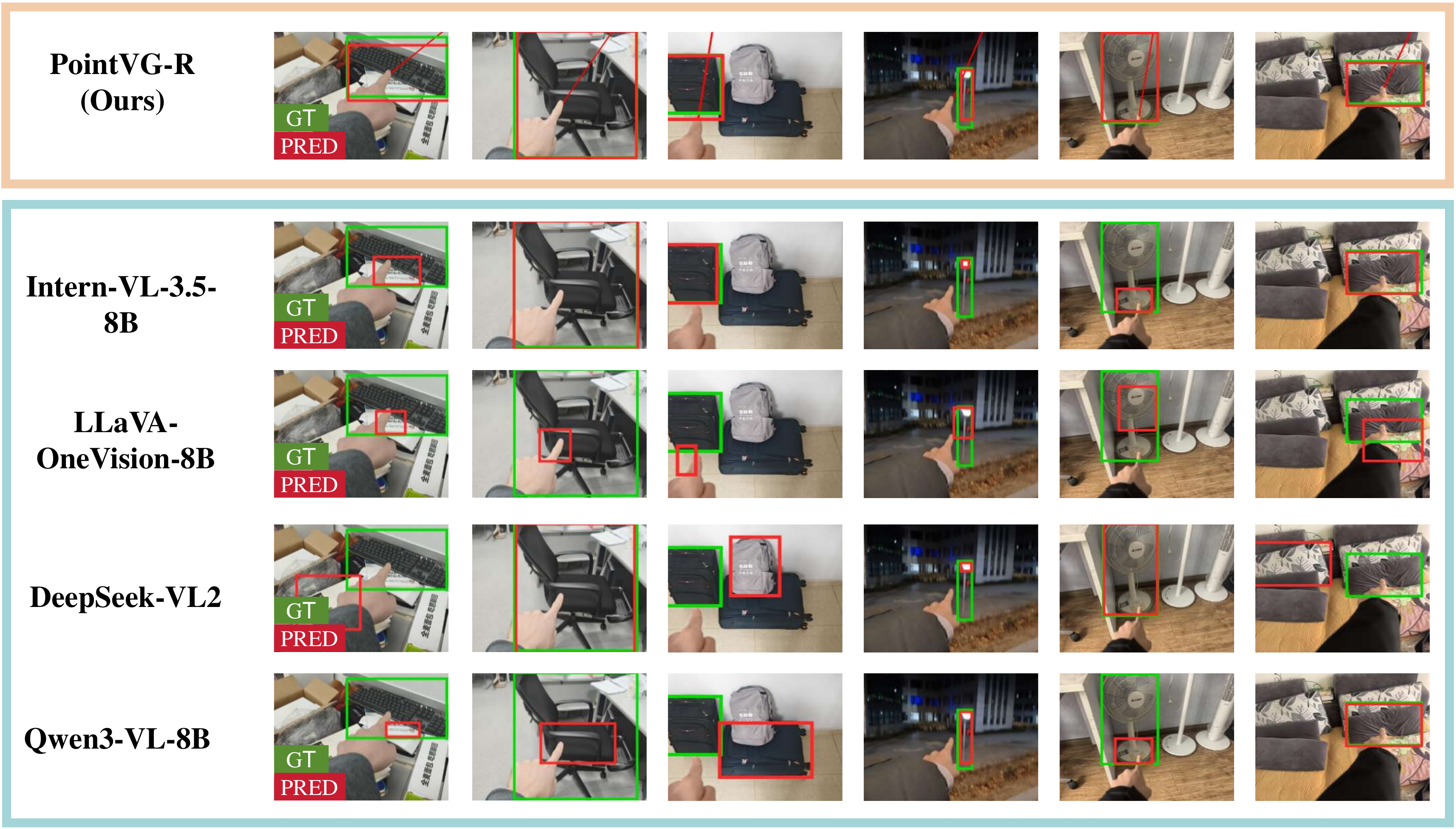}
    \caption{Comparative analysis of PointVG-R and baseline models on the EgoPoint-CoT.}
    \label{fig:figure5}
\end{figure*}
\subsection{Adaptive Reward Strategy}

During training, we observe that not all rollout groups provide equally informative learning signals. 
In some groups, the trajectory scores are concentrated within a consistently high or low range, which provides limited guidance for policy improvement. 
In contrast, groups with larger score variance indicate stronger relative preference among sampled trajectories and therefore offer more informative supervision. 
Motivated by this observation, we introduce an adaptive reward reweighting strategy that scales the contribution of each sample according to the variance of its rollout group.

Given a batch of trajectories, we first compute the sequence-level score $R_i$ for each sample $i$ by summing the token-level rewards penalized by the KL divergence \cite{kl1_kullback1951information}:
\begin{equation}
R_i = \sum_{t=1}^{|\mathcal{T}_i|} \left( s_{i,t} - \beta \mathrm{KL}_{i,t} \right),
\end{equation}
where $s_{i,t}$ is the environment reward at step $t$, and $\mathrm{KL}_{i,t} = \log \pi_{\theta_{\text{old}}}(a_{i,t} \mid x_i, a_{i,<t}) - \log \pi_{\text{ref}}(a_{i,t} \mid x_i, a_{i,<t})$ serves as a token-level penalty to prevent large deviations from the reference model.

Samples are grouped by their prompt identifier. For each group $G$ containing $|G|$ samples, we compute the mean $\mu_G$ and estimate the reward variance $\sigma_G^2$:
\begin{equation}
\mu_G = \frac{1}{|G|} \sum_{j \in G} R_j, \quad 
\sigma_G^2 = \frac{1}{|G|} \sum_{j \in G} (R_j - \mu_G)^2.
\end{equation}

To stabilize scaling across training steps, we maintain a running estimate of the global variance using an exponential moving average:
\begin{equation}
\sigma_{\text{global}}^2 \leftarrow 
\alpha \sigma_{\text{global}}^2 +
(1-\alpha)\sigma_{\text{batch}}^2,
\end{equation}
where $\sigma_{\text{batch}}^2$ is the mean variance across groups in the current batch and $\alpha$ is a momentum coefficient.

We then compute an importance weight $w_i$ for each sample based on the ratio between its group variance and the global variance:
\begin{equation}
w_i = \sqrt{\frac{\sigma_{G}^2}{\sigma_{\text{global}}^2 + \epsilon}},
\end{equation}
where $\epsilon$ is a small constant for numerical stability.

This adaptive strategy emphasizes samples from high-variance groups while down-weighting more stable ones, improving training stability and exploration efficiency.

We incorporate the proposed variance-based importance weighting into the GRPO objective by modulating the normalized advantages. The adjusted advantage for trajectory $\mathcal{T}_i$ is defined as:
\begin{equation}
\tilde{A}_i = \mathrm{clip}(w_i, w_{\min}, w_{\max}) \frac{R_i - \mu_G}{\sigma_G + \epsilon}.
\end{equation}

The resulting objective is formulated at the token level as:
\begin{flalign}
& \begin{aligned}
\mathcal{L}_{\text{GRPO}}(\theta)
&= - \mathbb{E}_{\mathcal{X}, \{R_i\}}
\Bigg[
\frac{1}{|G|} \sum_{i \in G} \sum_{t=1}^{|R_i|}
\min\Big(\rho_{i,t} \tilde{A}_i,\, \\
&\qquad \qquad \qquad \quad \mathrm{clip}(\rho_{i,t}, 1-\epsilon_c, 1+\epsilon_c) \tilde{A}_i\Big)
\Bigg],
\end{aligned} &
\end{flalign}
where $\rho_{i,t} = \frac{\pi_\theta(a_{i,t} \mid x_i, a_{i,<t})}{\pi_{\theta_{\text{old}}}(a_{i,t} \mid x_i, a_{i,<t})}$ denotes the policy ratio for token $t$ in trajectory $\mathcal{T}_i$. 




The clipping on $\rho_{i,t}$ stabilizes policy updates, while clipping $w_i$ prevents excessively large weights from outlier group variances, improving robustness. Introducing $w_i$ into the advantage explicitly modulates each trajectory’s contribution based on rollout group variance. High-variance groups are up-weighted due to stronger preference signals, whereas low-variance groups are down-weighted for their limited informativeness.

Importantly, $w_i$ is independent of the policy ratio $\rho_{i,t}$: $\rho_{i,t}$ captures policy shift, while $w_i$ reflects trajectory informativeness.Their combination gives an updated term:  $w_i \rho_{i,t} \frac{R_i - \mu_G}{\sigma_G + \epsilon}$ (with clipping), improving both optimization efficiency and training stability. This design allows GRPO to better exploit heterogeneous rollout quality without increasing gradient variance.

\vspace{-10pt}
\section{EgoPoint-CoT Dataset}

To support the explicit spatial reasoning required during the SFT cold-start phase, we construct EgoPoint-CoT, whose annotation pipeline is illustrated in Figure~\ref{fig:figure4}.
comprising approximately 15K images (split into train, validation, and test sets at a 7:2:1 ratio). Built upon EgoPoint-Ground \cite{egopoint_li2026languagegroundingreferringexpressions}, this dataset provides complete $(R^*, b^*)$ supervision signals that strictly align with our four-stage reasoning trajectory. Specifically, we extract physical anchors ($r_1$ and $r_2$) via MMPose \cite{mmpose2020} coupled with expert cross-validation, and incorporate target semantics ($r_3$) and ground-truth boxes ($r_4$) to generate formatted reasoning sequences via LLMs, followed by manual verification. Furthermore, to suppress model hallucinations, we explicitly inject 1,052 negative samples (images without visible pointing hands) paired with blocking reasoning trajectories (e.g., ``No hand detected''). By blending real-world smart-glass captures with synthetic renderings, EgoPoint-CoT ensures both rich scene diversity and robust distributional consistency.
\section{Experiments}

\begin{table*}[!t]
\centering
\caption{Main results on the EgoPoint-CoT benchmark. We compare our PointVG-R against a wide range of state-of-the-art VLMs and grounding models across zero-shot, fine-tuned, and CoT-based settings. By integrating structured V-CoT reasoning with Adaptive GRPO, PointVG-R substantially surpasses existing methods, effectively bridging the divide between current model capabilities and human-level spatial reasoning.}
\label{tab:main_results}
\resizebox{0.7\textwidth}{!}{
\begin{tabular}{l l c cccc}
\toprule
\textbf{Training Paradigm} & \textbf{Model} & \textbf{Backbone} & \textbf{P@0.3}$\uparrow$ & \textbf{P@0.5}$\uparrow$ & \textbf{P@0.7}$\uparrow$ & \textbf{mIoU}$\uparrow$ \\
\midrule
\textit{Upper Bound} & Human Study & -- & 0.9398 & 0.8850 & 0.8218 & 0.8265 \\
\midrule
\rowcolor[gray]{0.9} \multicolumn{7}{l}{\textbf{Zero-shot Methods}} \\
\multirow{10}{*}{N/A} 
& DeepSeek-OCR \cite{wei2025deepseek} & -- & 0.3853 & 0.3207 & 0.2734 & 0.3367 \\
& DeepSeek-VL2-Small \cite{wu2024deepseek} & -- & 0.4077 & 0.3738 & 0.3351 & 0.3619 \\
& Ferret-v1 \cite{you2023ferret} & 7B & 0.3489 & 0.2954 & 0.2428 & 0.2956 \\
& Ferret-v1 \cite{you2023ferret} & 13B & 0.3059 & 0.2471 & 0.1998 & 0.2634 \\
& Florence2-large \cite{xiao2024florence} & -- & 0.1079 & 0.0869 & 0.0791 & 0.1086 \\
& InternVL-v3.5 \cite{wang2025internvl3} & 8B & 0.4790 & 0.4264 & 0.3466 & 0.3989 \\
& LLaVA-OneVision \cite{an2025llava} & 8B & 0.5258 & 0.3987 & 0.2562 & 0.4234 \\
& LLaVA-v1.5 \cite{liu2023visual} & 13B & 0.3810 & 0.2715 & 0.1138 & 0.2642 \\
& Qwen2.5-VL \cite{wang2024qwen2} & 7B & 0.5421 & 0.4976 & 0.4293 & 0.4549 \\
& Qwen3-VL \cite{bai2025qwen3} & 8B & 0.4814 & 0.4446 & 0.3896 & 0.4298 \\
\midrule
\rowcolor[gray]{0.9} \multicolumn{7}{l}{\textbf{Fine-tuned Baselines (Standard SFT)}} \\
\multirow{4}{*}{Non Thinking SFT}
& LLaVA-v1.5 \cite{liu2023visual} & 13B &0.6978 &0.5913 &0.3618 &0.5165 \\
& InternVL-3.5 \cite{wang2025internvl3} & 8B &0.6802 &0.6319 &0.5506 &0.5867 \\
& Qwen2.5-VL \cite{wang2024qwen2} & 7B &0.6873 & 0.6362 &0.5674 &0.5984 \\
& Qwen3-VL \cite{bai2025qwen3} & 8B &0.6266 &0.5932 &0.5387 &0.5734 \\
\midrule
\rowcolor[gray]{0.9} \multicolumn{7}{l}{\textbf{Visual Chain-of-Thought (Cold Start)}} \\
\multirow{3}{*}{SFT-CoT ($r_1 \sim r_4$)}
& LLaVA-v1.5 \cite{liu2023visual} & 13B & 0.6934 & 0.5778 & 0.3517 & 0.5225 \\
& InternVL-3.5 \cite{wang2025internvl3} & 8B &0.6089 &0.5650 &0.4947 & 0.5168 \\
& Qwen2.5-VL \cite{wang2024qwen2} & 7B & 0.8025 & 0.6935 & 0.5449 & 0.6453  \\
& Qwen3-VL \cite{bai2025qwen3} & 8B &0.7347 &0.7021 & 0.6438 &0.6688\\
\midrule
\rowcolor[gray]{0.9} \multicolumn{7}{l}{\textbf{Ours: PointVG-R (V-CoT + Adaptive GRPO)}} \\
{\textbf{V-CoT + RL}}
& \textbf{PointVG-R (Qwen2.5)} & 7B &\textbf{0.8805} 
& \textbf{0.8360} 
& \textbf{0.7452} 
& \textbf{0.7570} \\
\bottomrule
\end{tabular}
}
\end{table*}

\subsection{Main Results}
We have conducted a comprehensive evaluation of PointVG-R against SOTA MLLMs on the EgoPoint-CoT test set. Table \ref{tab:main_results} presents the quantitative results across various training paradigms, from which we derive the following key insights:

\noindent\textbf{Cognitive Vulnerability of Traditional Models.} Baseline models exhibit significant cognitive vulnerabilities in pointing-based visual grounding. Zero-shot results show consistently poor performance; even high-performing models like Qwen2.5-VL (7B) and LLaVA-OneVision (8B) struggle, with $mIoU$ scores stagnating between $0.42$ and $0.45$. This highlights the inability of general-purpose models to decipher complex gestural spatial relations without specialized reasoning. Qualitative comparisons in Figure~\ref{fig:figure5} further corroborate these findings, showing that PointVG-R produces more spatially consistent and semantically accurate localizations than competing baselines.

Comparing standard SFT with our V-CoT cold-start setting reveals that structured intermediate supervision can strengthen geometric perception, although the effect varies across model families. Specifically, on the Qwen2.5-VL backbone, the V-CoT paradigm with explicit intermediate supervision improves $mIoU$ from 0.5984 to 0.6453 ($\uparrow$ 4.7 points), and Qwen3-VL also achieves a clear gain from 0.5734 to 0.6688 ($\uparrow$ 9.5 points).
In contrast, LLaVA-v1.5 shows only marginal improvement under CoT supervision, with $mIoU$ changing from $0.5165$ to $0.5225$, while InternVL-3.5 drops from $0.5867$ to $0.5168$ ($\downarrow 7.0$ points). This indicates that the efficacy of structured reasoning depends on the inherent capacity of the model backbone. By emulating human cognitive processes, we enable compatible models to transcend simple feature matching and achieve sophisticated visual logical reasoning.

\noindent\textbf{Performance Gains from Adaptive RL.} PointVG-R (Ours) achieves SOTA performance across all metrics. By integrating the GRPO algorithm with our adaptive importance weighting strategy during the second stage, PointVG-R reaches an $mIoU$ of $0.7570$ on the Qwen2.5 (7B) backbone—representing a \textbf{$\textbf{8.82}$} points improvement over the strongest cold start baseline (\textbf{$0.6688 \rightarrow 0.7570$}). This substantial gain validates that our group-variance-based weighting dynamically optimizes learning signals, effectively addressing the challenge of inconsistent reward quality and further unleashing the model's reasoning potential.
\subsection{Ablation Study}
\noindent \textbf{Experiment I: Ablation on Intermediate V-CoT Supervision.} \\
We conducted a progressive ablation study on the intermediate reasoning components ($f_1 \sim f_3$) to evaluate the effect of supervising explicit geometric deduction stages during the GRPO phase. As reported in Table~\ref{tab:paradigm_ablation}, intermediate rewards are essential for establishing logical consistency, with the full model achieving overall gains of +4.07, +4.40, +5.88, and +4.03 points on $P@0.3$, $P@0.5$, $P@0.7$, and $mIoU$, respectively, over the baseline. First, incorporating hand localization supervision ($f_1$) significantly stabilizes the grounding process. Compared with the ``w/o inter.'' variant, adding the hand prior improves $P@0.3$ from $0.8398$ to $0.8814$ ($\uparrow 4.16$ points) and $mIoU$ from $0.7167$ to $0.7463$ ($\uparrow 2.96$ points). This substantial improvement across all metrics---including gains of +4.11 on $P@0.5$ and +3.53 on $P@0.7$---confirms that the hand serves as a critical spatial anchor. Without this explicit supervision, the model must implicitly infer the interaction source, which increases spatial ambiguity and leads to suboptimal convergence. Second, introducing ray and keypoint supervision ($f_2$) primarily improves fine-grained spatial alignment. Although $P@0.3$ remains nearly saturated ($0.8814 \rightarrow 0.8805$), the inclusion of $f_2$ yields a clear improvement on the more stringent $P@0.7$ metric, increasing it from $0.7217$ to $0.7452$ ($\uparrow 2.35$ points), while further raising $mIoU$ to $0.7570$ ($\uparrow 1.07$ points).

\vspace{10pt}
\begin{table*}[!t]
\centering
\caption{Ablation study on the structured V-CoT reasoning paradigm. We evaluate the incremental contribution of each physical modeling component ($f_1$ to $f_3$) and the impact of the image-ray re-encoding module. The results validate the necessity of each stage in forming a deterministic reasoning chain for precise grounding.}
\label{tab:paradigm_ablation}

\setlength{\tabcolsep}{4pt}
\renewcommand{\arraystretch}{1.08}
\footnotesize

\begin{tabular}{@{}lccccccc@{}}
\toprule
\multirow{2}{*}{\textbf{Method / Paradigm}} 
& \multicolumn{3}{c}{\textbf{V-CoT Components}} 
& \multicolumn{4}{c}{\textbf{Performance Metrics}} \\
\cmidrule(lr){2-4} \cmidrule(lr){5-8}
& \textbf{Hand ($f_1$)} 
& \textbf{Ray \& Point ($f_2$)} 
& \textbf{Box ($f_3$)} 
& \textbf{P@0.3}$\uparrow$ 
& \textbf{P@0.5}$\uparrow$ 
& \textbf{P@0.7}$\uparrow$ 
& \textbf{mIoU}$\uparrow$ \\
\midrule

\rowcolor[gray]{0.85}
\multicolumn{8}{l}{\textbf{Baseline: Qwen2.5-VL-7B}} \\

Visual CoT (w/o inter.) 
& \redx 
& \redx 
& \greencheck
& 0.8398
& 0.7920
& 0.6864
& 0.7167 \\

Visual CoT ($+r_1$) 
& \greencheck 
& \redx
& \greencheck 
& 0.8814
& 0.8331
& 0.7217
& 0.7463 \\

\textbf{Ours PointVG-R (Full)} 
& \greencheck 
& \greencheck 
& \greencheck 
& \textbf{0.8805} 
& \textbf{0.8360} 
& \textbf{0.7452} 
& \textbf{0.7570} \\
\bottomrule
\end{tabular}
\end{table*}

\begin{table}[h]
\centering
\caption{Ablation study on group statistics for reward normalization in GRPO}
\label{tab:group_stat}
\resizebox{\columnwidth}{!}{ 
\begin{tabular}{l c c c c}
\toprule
 \textbf{Statistic Setting} & \textbf{P@0.3} & \textbf{P@0.5} & \textbf{P@0.7} & \textbf{mIoU} \\
\midrule
None (Standard GRPO) &0.8240  &0.7773  &0.6921  &0.7184  \\
Reward Entropy &0.7490  &0.6716 &0.5822 &0.6373 \\
Top-Bottom Reward Gap &0.4010  &0.3714 &0.3303 &0.3618 \\
Standard Deviation &0.8436 &0.7921 &0.6916 &0.7212 \\
Group Variance (Ours) &\textbf{0.8805} 
& \textbf{0.8360} 
& \textbf{0.7452} 
& \textbf{0.7570} \\
\bottomrule
\end{tabular}
} 
\end{table}

\begin{table}[h]
\centering
\caption{Ablation study on importance mapping functions for reward scaling.}
\label{tab:mappingFunc_stat}
\resizebox{\columnwidth}{!}{
\begin{tabular}{l c c c c}
\toprule
\textbf{Mapping Function} & \textbf{P@0.3} & \textbf{P@0.5} & \textbf{P@0.7} & \textbf{mIoU} \\
\midrule
Variance + Linear Scaling & 0.7203 &0.6644  &0.5793  & 0.6017 \\
Variance + Power Scaling &0.8369  &0.7868 &0.6826 &0.7112  \\
Variance + Log Scaling &0.8485  &0.8006  &0.7060 &0.7269 \\

Variance + Sqrt Scaling (Ours) & \textbf{0.8805} 
& \textbf{0.8360} 
& \textbf{0.7452} 
& \textbf{0.7570}  \\
\bottomrule
\end{tabular}
}
\end{table}
\noindent \textbf{Experiment II: Ablation on Group Statistic for Importance Estimation.} \\
\vspace{-5pt}
To investigate the optimal group-level statistic for guiding adaptive reward weighting, we conducted a systematic ablation by comparing Group Variance ($\sigma_G^2$) against vanilla GRPO, reward entropy, top-bottom reward gap, and standard deviation, while keeping all other components constant. As summarized in Table~\ref{tab:group_stat}, Group Variance achieves superior performance across all metrics, reaching \textbf{0.8805/0.8360/0.7452} on $P@0.3/P@0.5/P@0.7$ and an $mIoU$ of 0.7570. Compared to vanilla GRPO, it yields consistent precision improvements exceeding 5.0 points and a 3.86 absolute gain in $mIoU$, validating the critical role of modeling intra-group rollout heterogeneity in stabilizing policy updates for complex spatial reasoning. Comparative analysis further reveals that Group Variance significantly outperforms first-order and information-theoretic measures. Specifically, the top-bottom reward gap exhibits the poorest performance, with an $mIoU$ of 0.3618, due to its inherent sensitivity to sampling outliers, while reward entropy fails to directly reflect the geometric alignment quality essential for V-CoT trajectories. Notably, Group Variance outperforms standard deviation by 3.58 points in $mIoU$, suggesting that its second-order nature provides a more discriminative scaling signal. This enables the model to effectively prioritize high-fidelity reasoning paths while suppressing noise from suboptimal rollouts. In summary, Group Variance serves as a robust proxy for rollout quality, substantially enhancing policy optimization efficiency.

\vspace{10pt}
\noindent \textbf{Experiment III: Ablation on Importance Mapping Function.} \\




Upon establishing group variance as the core statistic, we further investigate the functional form used to map it into the final importance weight $w_G$. Maintaining constant variance terms, we evaluate linear, power, and logarithmic scaling against our proposed square-root scaling ($w_G \propto \sqrt{\sigma_G^2}$). As summarized in Table~\ref{tab:mappingFunc_stat}, square-root scaling achieves the most robust performance with an $mIoU$ of 0.7570, significantly outperforming the linear baseline (0.6017). These results indicate thato tackle pointing tasks with complex spatial structures effectively a stable optimization signal. While linear scaling proves overly aggressive—potentially inducing gradient instability during training—logarithmic scaling (0.7269) suffers from over-compression of inter-group heterogeneity, leading to insufficient sensitivity toward high-fidelity reasoning paths. In contrast, square-root scaling implements a sub-linear growth pattern that effectively captures group-level distinctions while ensuring smooth weight transitions. This approach strikes an optimal balance between responsiveness and training stability, ultimately yielding superior performance.



\vspace{-5pt}
\section{Conclusion}
MLLM is specifically designed for pointing gesture visual grounding through Visual Chain-of-Thought (V-CoT) reasoning. PointVG-R emulates human iterative cognitive processes to effectively tackle pointing tasks with complex spatial structures. We introduce a two-stage training framework: the first stage utilizes "cold-start" supervised fine-tuning (SFT) to establish a "thinking-with-images" geometric reasoning pipeline; the second stage leverages the GRPO algorithm to further bolster reasoning capabilities. Furthermore, we develop a group-variance-based adaptive importance weighting strategy, which enhances overall performance and convergence stability by dynamically prioritizing sample groups with pronounced relative preferences. Extensive evaluations demonstrate that PointVG-R achieves a $\textbf{15.86\%}$ performance gain over baseline methods, while detailed ablation studies confirm the efficacy of our reasoning chain and innovative modules. Ultimately, PointVG-R exhibits superior versatility and robustness in handling human-centric spatial interaction tasks.

\noindent\textbf{Limitation}: The iterative V-CoT paradigm employed in PointVG-R requires sequential token generation for multiple intermediate reasoning steps (e.g., $r_1$ to $r_4$). Unlike traditional direct-regression approaches, this autoregressive process introduces non-negligible inference latency. For real-time applications on wearable devices such as AR glasses, maintaining complex logical reasoning while reducing token overhead or accelerating inference remains a critical challenge for future work.


\newpage
\appendix

\section{Related Work}
\subsection{Visual Grounding and First-Person Pointing Understanding}
The visual grounding task has evolved from earlier referring expression comprehension (RefCOCO \cite{yu2016refcoco}\cite{lin2014microsoftcoco}, RefCOCO+ \cite{mao2016refcoco+}, RefCOCOg \cite{nagaraja2016refcocog}) to open-vocabulary detection (GLIP \cite{li2022grounded_GLIP}, Grounding DINO \cite{liu2024grounding_groundingdino}, YoloWorld \cite{cheng2024yolo_yoloworld}) and integration with MLLMs (LLaVA-1.6 \cite{liu2024llavanext_llava}\cite{liu2024improved_llava}\cite{liu2024improved_llava}, Qwen-VL \cite{wang2024qwen2_Qwenvl}, Deepseek-OCR \cite{wei2025deepseek_deepseekocr}). While these methods perform well in general scenarios, their designs typically assume static, third-person images as input, making them less effective for handling dynamic interactions from a first-person perspective. In recent years, Ego4D \cite{grauman2022Ego4d}, EgoExo4D \cite{grauman2024ego_egoexo4d}, and their extensions (EgoObjects \cite{zhu2023egoobjects}) have propelled research in egocentric vision, fostering tasks such as hand detection (HandNet), active object detection, and gesture-object association modeling (Pointing Grounding in Ego4D \cite{grauman2022Ego4d}, EgoPoint \cite{darkhalil2025egopoints}). Recent works like EgoVLP \cite{pramanick2023egovlpv2_egovlp} and EgoVLA \cite{yang2025egovla_egovla} attempt to align language instructions with first-person vision but still simplify the pointing task to end-to-end bounding box regression, lacking explicit modeling of the "gesture-direction-target" geometric chain.

\subsection{Multimodal Chain-of-Thought and Structured Reasoning}
Chain-of-Thought (CoT) has expanded from the pure text domain \cite{wei2022chain_cot} to visual reasoning (Think with image \cite{think_with_image1,wei2022chain_cot,think_with_image3,think_with_image4,think_with_image5,think_with_image6,think_with_image7,think_with_image8,think_with_image9,think_with_image10,think_with_image11,think_with_image12}. VisProg \cite{gupta2023visual_visprog} and ViperGPT \cite{suris2023vipergpt} achieve interpretable reasoning through program synthesis; Program-of-Thought \cite{chen2022program_pot} introduce external tools to enhance symbolic manipulation capabilities. Recently, end-to-end MLLMs have also begun to support implicit CoT, such as LLaVA-CoT \cite{xu2025llava_llavacot} and Qwen-VL-CoT \cite{wang2024qwen2_Qwenvl}. However, their reasoning steps remain primarily text-based, without enforcing alignment between intermediate representations and visual geometric structure. Furthermore, some works have explored structured outputs for spatial reasoning, such as GeoQA+ \cite{cao2022augmented_geoqa+} (geometry problem solving) and Diagram-Guided Reasoning \cite{wang2024cog_Diagram-Guided_Reasoning} (diagram understanding). Yet, these methods rely on domain-specific priors, making it difficult to generalize to open-domain first-person pointing tasks.

\subsection{Spatial Intelligence and Reinforcement Learning in MLLMs}
Despite progress in perception tasks \cite{caffagni2024revolution_MLLM}, the spatial reasoning capabilities of MLLMs remain questionable. Studies indicate systematic deficiencies in models' understanding of coordinates, directional inference, and topological relationship modeling \cite{anand2025cube_MLLM1}\cite{o2024metric_MLLM2}\cite{yin2025spatial_MLLM3}\cite{li2024reframing_MLLM4}. To enhance these capabilities, researchers have attempted to introduce synthetic data (SpatialVLM \cite{chen2024spatialvlm_SpatialVLM}), neuro-symbolic interfaces (Neuro-Symbolic VQA \cite{berlot2021neuro_Neuro-Symbolic_VQA}), or geometric pre-training (GeoFormer \cite{zhao2023geoformer_GeoFormer}). Concurrently, reinforcement learning with verifiable rewards (RLVR) \cite{RLVR1}\cite{RLVR2}\cite{RLVR3}\cite{RLVR4}\cite{RLVR5}\cite{RLVR6} is widely employed to improve the reasoning performance of MLLMs (e.g., o1 \cite{jaech2024openai_o1}, DeepSeek-R1 \cite{guo2025deepseek_deepseekr1}). This work demonstrates that optimization based on verifiable rewards can effectively enhance sampling efficiency. However, the focus has primarily been on mathematical/coding domains and has not yet been systematically applied to visual grounding tasks with strong geometric constraints.

\section{Experiment Details}
\subsection{Experimental Settings}

\noindent\textbf{Baselines and Evaluation Metrics.}
We compare our method against both large-model-based grounding approaches and general-purpose vision-language models, including Qwen-VL (2.5/3B--8B), LLaVA (v1.5/OneVision, 7B--13B), InternVL-3.5 (4B/8B), Ferret (7B/13B), DeepSeek (VL2-Small/OCR). All baselines use their official open-source implementations and generate target bounding boxes via native autoregressive decoding under the same input format (image + pointing query), without external tools or post-processing. Evaluation is conducted using Precision@0.3, Precision@0.5, Precision@0.7, and mean IoU (mIoU) as metrics.

Given a dataset of $N$ samples, $\text{Precision}@K$ evaluates the proportion of predictions that exceed a specific IoU threshold $K$:
\[
\text{Precision}@K = \frac{1}{N} \sum_{i=1}^{N} \mathbb{I}\left( \mathrm{IoU}(\hat{b}^{(i)}, b^{*(i)}) > K \right),
\]
and mIoU is computed as the average intersection-over-union across all samples:
\[
\text{mIoU} = \frac{1}{N} \sum_{i=1}^{N} \mathrm{IoU}(\hat{b}^{(i)}, b^{*(i)}),
\]
where $\mathbb{I}(\cdot)$ is the indicator function, $\hat{b}^{(i)}$ represents the predicted bounding box, $b^{*(i)}$ denotes the ground-truth bounding box, and $\mathrm{IoU}(\hat{b}, b^*) = \frac{|\hat{b} \cap b^*|}{|\hat{b} \cup b^*|}$.

\noindent\textbf{Training and inference Details.}
We instantiate PointVG-R on top of \textbf{Qwen2.5-VL-7B} and optimize it with the two-stage training protocol introduced in main paper, a cold-start CoT-SFT stage followed by an adaptive GRPO stage. In the \textbf{SFT stage}, we use the \emph{entire train split} of EgoPoint-CoT for supervised learning. To improve training efficiency while preserving the structured reasoning capability established by cold-start supervision, the subsequent {GRPO stage} uses a \emph{fixed random 30\% subset} of the train split. This subset is sampled once before RL training starts and is kept unchanged throughout all epochs, i.e., we do \emph{not} re-sample the 30\% subset at the beginning of each epoch.

For the cold-start phase, we adopt {LoRA} for parameter-efficient fine-tuning. The SFT stage is trained for 3 epochs with LoRA rank 8 and LoRA alpha 16. The learning rate is set to $1\times10^{-4}$, and the global batch size is 384. In addition to the standard LoRA trainable parameters, we further unfreeze the {token embedding layer} and the {lm\_head}, since our structured V-CoT formulation introduces additional special tokens for point-, caption-, and tool-related syntax. This stage is used to inject a stable geometric reasoning prior and to initialize the model with a well-formed, structured decoding policy before reinforcement learning.

In the second stage, we perform {full-parameter GRPO training} for 7 epochs. The actor optimizer is AdamW with learning rate $1\times10^{-5}$, weight decay 0.01, and $\beta=(0.9, 0.999)$. We use a cosine learning-rate scheduler with a warmup ratio 0.1 and a minimum learning-rate ratio of 0.2. For policy updates, the actor global batch size is 128. We set the maximum gradient norm to 0.5 and use one PPO epoch per update. The policy clipping parameters are set to clip-ratio-low $=0.2$, clip-ratio-high $=0.3$, and dual-clip $=3.0$.

For rollout generation, we use a {vLLM}-based hybrid engine with rollout batch size 256. During training, we sample $n=5$ responses for each prompt in each rollout step, while during validation, we use $n=3$ sampled generations per instance. The sampling temperature is set to 0.9, with top-$p=0.95$ and top-$k=-1$. Since our V-CoT reasoning includes tool invocation and image-ray-based intermediate reasoning, we enable {multi-turn rollout} with at most 2 iterations. The maximum prompt length and maximum response length are both set to 1024, and the maximum generation length per turn is also 1024. During GRPO training, we additionally enable \textbf{Flash Attention} to improve memory efficiency and training throughput.

The reward follows the structured design in Sec.~3.3. Specifically, the base reward is defined as
\[
R_{\text{base}} = R_{\text{hand}} + R_{\text{ray}} + R_{\text{kpt}} + \lambda_1 R_{\text{iou}} + \lambda_2 R_{\text{fmt}},
\]
where the coefficients are set to $\lambda_1=5$ and $\lambda_2=2$. We keep the repetition penalty and negative-sample penalty described in Sec.~3.3 unchanged during training. For KL control, we enable KL loss with coefficient 0.05, adopt the \textbf{low\_var\_kl} penalty, and use an adaptive KL controller with target KL 0.08 and horizon 10000.

For the proposed adaptive reward strategy in Sec.~3.4, we use \textbf{group variance} as the rollout-quality statistic and adopt the \textbf{sqrt-ratio} mapping to compute the importance weight. The global variance is initialized as 0.1 and updated with an exponential moving average whose momentum coefficient is $\alpha=0.95$. The numerical stabilizer is set to $10^{-6}$. The resulting importance weight is clipped to the interval $[0.3,\,3.0]$ before being applied to the normalized group advantage. We do not apply additional normalization to the importance weights.

At the data-processing level, we shuffle the GRPO training subset and set the image resolution range to [262144,\,1048576] pixels and filter out overlong prompts before training. During validation, the batch size is set to 256. We perform validation once every epoch and also evaluate once before training starts. For validation-time generation, we use a more conservative decoding setup with temperature 0.6, top-$p$ 0.95, and 3 sampled generations per instance.

All experiments are conducted on a single node with \textbf{8 NVIDIA H100 GPUs}.
\subsection{Details for Ablation Study}

All ablation experiments follow the same training protocol as the standard model in the main paper. Concretely, all variants are initialized with the same backbone and structured Visual Chain-of-Thought (V-CoT) framework, use 100\% of the SFT training data for cold-start supervised fine-tuning, and further use 30\% of the training data for RL optimization. All other optimization and model configurations are kept identical to those used in the main experiments, so that the comparisons isolate the effect of the ablated component.

\subsubsection{Ablation on the Structured V-CoT Reasoning Paradigm.}

To verify the necessity of explicit intermediate supervision in the proposed reasoning framework, we perform a progressive ablation study by gradually adding reward supervision signals into the RL stage. Following the four-stage reasoning process described in the main paper, the final object box prediction is always retained, while the intermediate physical cues are introduced step by step. Specifically, we start from a reduced setting that only supervises the final target box, then add hand localization supervision, and finally introduce ray and pointing keypoint supervision. The corresponding quantitative results are reported in Table~2 of the main paper.

This design directly evaluates whether structured geometric supervision improves policy learning beyond final-box-only optimization. As shown in Table~2 of the main paper, the model with only final box supervision already provides a reasonable grounding baseline, but adding explicit hand supervision significantly improves all metrics, indicating that the pointing hand serves as a critical spatial anchor. Further introducing ray and point supervision leads to additional gains, especially on stricter localization metrics such as P@0.7 and mIoU. These results support the central claim of our method: precise pointing localization benefits from decomposing the task into interpretable geometric sub-problems rather than optimizing only the final grounding output.

\subsubsection{Ablation on Group Statistics for Importance Estimation.}

The main paper adopts group variance as the statistic used to estimate the informativeness of each rollout group. To verify whether this choice is indeed the most effective, we compare it against several alternative group-level statistics under the same RL setting. For one rollout group containing $K$ sampled responses, let the reward set be
\[
\mathbf{r}=\{r_1,r_2,\dots,r_K\},
\qquad
\bar{r}=\frac{1}{K}\sum_{i=1}^{K} r_i.
\]
Based on this reward set, we consider the following candidate statistics.

\textbf{(1) Variance}
\[
S_{\mathrm{var}}(\mathbf{r})
=
\frac{1}{K}\sum_{i=1}^{K}(r_i-\bar{r})^2.
\]

\textbf{(2) Standard deviation}
\[
S_{\mathrm{std}}(\mathbf{r})
=
\sqrt{
\frac{1}{K}\sum_{i=1}^{K}(r_i-\bar{r})^2
}.
\]

\textbf{(3) Top-bottom gap}
\[
S_{\mathrm{gap}}(\mathbf{r})
=
\max_{1\le i\le K} r_i-\min_{1\le i\le K} r_i.
\]

\textbf{(4) Reward entropy}
First, we partition the reward range
\[
[r_{\min}, r_{\max}],
\qquad
r_{\min}=\min_i r_i,\;\;
r_{\max}=\max_i r_i
\]
into $B$ histogram bins, and denote by $p_b$ the normalized frequency of the $b$-th bin, where
\[
p_b\ge 0,
\qquad
\sum_{b=1}^{B} p_b = 1.
\]
The entropy statistic is then defined as
\[
S_{\mathrm{ent}}(\mathbf{r})
=
-\sum_{b=1}^{B} p_b \log(p_b+\varepsilon),
\]
where $\varepsilon>0$ is a small constant for numerical stability.

For degenerate groups, we follow a stable implementation rule. When $K\le 1$, the statistic is set to zero:
\[
S(\mathbf{r})=0.
\]
For the entropy-based statistic, when $r_{\max}-r_{\min}\le \varepsilon$, we also set
\[
S_{\mathrm{ent}}(\mathbf{r})=0.
\]

The corresponding quantitative comparison is reported in Table~3 of the main paper. As shown there, group variance achieves the best performance across all metrics. Compared with standard GRPO without adaptive statistic-based weighting, variance-based scaling yields clear gains on both localization precision and mIoU. In contrast, top-bottom gap is highly sensitive to outlier rewards, while entropy does not directly capture the relative geometric discriminability among sampled responses. Standard deviation performs substantially better than these alternatives, but still underperforms variance. These observations indicate that a second-order statistic over rollout rewards provides the most reliable signal for measuring the usefulness of a sampled group in our structured spatial reasoning setting.

\subsubsection{Ablation on Importance Mapping Functions for Reward Scaling.}
After selecting the group statistic, we further study how the statistic value should be mapped into the final importance weight for advantage rescaling. Let $s_g$ denote the statistic value of group $g$, computed by the selected statistic function. Each sample $i$ in group $g$ inherits the same statistic,
\[
s_i = s_g.
\]
Let the baseline statistic be denoted by $b$. In implementation, $b$ can be either the current batch average statistic plus a numerical constant, or a running global baseline statistic. Specifically, the batch baseline is
\[
b_{\mathrm{batch}}=\bar{s}_{\mathrm{batch}}+\varepsilon,
\qquad
\bar{s}_{\mathrm{batch}}=\frac{1}{G}\sum_{g=1}^{G} s_g,
\]
where $G$ is the number of rollout groups in the batch. When a global baseline is used, it is updated by exponential moving average:
\[
b_{\mathrm{global}}^{(t)}
=
\max\!\Bigl(
\varepsilon,\;
m\, b_{\mathrm{global}}^{(t-1)}
+
(1-m)\,\bar{s}_{\mathrm{batch}}^{(t)}
\Bigr),
\]
where $m\in[0,1)$ is the momentum coefficient. The actual baseline is then
\[
b=
\begin{cases}
b_{\mathrm{batch}}, & \text{if batch statistic is used},\\
b_{\mathrm{global}}, & \text{otherwise}.
\end{cases}
\]

Given the per-sample statistic $s_i$ and the baseline $b$, we evaluate the following candidate mappings.

\textbf{(1) Square-root ratio}
\[
w_i^{(\mathrm{sqrt})}
=
\sqrt{\frac{s_i}{b}}.
\]

\textbf{(2) Linear ratio}
\[
w_i^{(\mathrm{linear})}
=
\frac{s_i}{b}.
\]

\textbf{(3) Power ratio}
\[
w_i^{(\mathrm{power})}
=
\left(\frac{s_i}{b}\right)^{\gamma},
\]
where $\gamma>0$ controls the nonlinearity. In our implementation, the default choice is $\gamma=1.5$.

\textbf{(4) Log ratio}
\[
w_i^{(\mathrm{log})}
=
\log\!\left(1+\alpha\frac{s_i}{b}\right),
\]
where $\alpha>0$ is a scaling factor.

After computing the raw importance weights, we optionally apply clipping:
\[
\tilde{w}_i = \operatorname{clip}(w_i, w_{\min}, w_{\max}).
\]
The final weight becomes
\[
A_i' = \tilde{w}_i A_i.
\]

The corresponding quantitative comparison is reported in Table~4 of the main paper. Among all mapping functions, square-root scaling yields the strongest overall results. Linear scaling is too aggressive and can amplify unstable groups excessively, while logarithmic scaling tends to over-compress the differences among groups. Power scaling provides a more moderate alternative, but still remains inferior to square-root scaling. Overall, these results suggest that sub-linear scaling offers the best balance between sensitivity to informative groups and robustness during RL optimization.

\subsubsection{Summary.}
Overall, the ablation experiments consistently validate the two key design choices of PointVG-R. First, the structured V-CoT reasoning paradigm benefits substantially from progressively injected intermediate supervision, confirming that explicit hand-, point-, and ray-level reasoning improves final grounding quality. Second, within adaptive RL optimization, group variance serves as the most effective statistic for capturing rollout informativeness, while square-root scaling provides the most stable mapping from group-level uncertainty to sample-level importance. Together, these findings support the effectiveness of the proposed structured reasoning and adaptive reward weighting design for precise pointing localization.

\subsection{Pseudo-code for the Adaptive Reward Strategy}
At each training iteration, a batch of prompt-image inputs is first sampled from the dataset. For each input $x$, the current policy generates a group of $G$ responses via multi-turn rollout. These responses are then evaluated using a task-specific reward function (EgoReward), producing a set of reward signals. Based on these rewards, group-relative advantages are computed using the GRPO advantage estimator.

To capture the variability of the generated responses, the algorithm computes the variance of rewards within each group as a group-level statistic. These statistics are averaged across the batch to obtain a batch-level estimate.

A key component of the method is adaptive advantage scaling. For each input, a scaling weight is computed based on the ratio between its group variance and a global running statistic. This weight is further clipped within predefined bounds to ensure stability, and then applied to rescale the corresponding advantages. This mechanism encourages the model to place more emphasis on samples with higher reward diversity.

The global statistic is updated using an exponential moving average (EMA), ensuring a smooth and stable estimate over time.

Finally, the policy parameters are updated using the GRPO objective with KL regularization against a reference policy $\pi_{\mathrm{ref}}$, balancing exploration and stability.

\begin{algorithm}[t]
\caption{PointVG-R Training}
\label{alg:ego_cot_ab}
\begin{algorithmic}[1]
\Require policy $\pi_\theta$, reference policy $\pi_{\mathrm{ref}}$, dataset $D$
\Require group size $G$, momentum $\beta$, initial global statistic $s_{\mathrm{g}}$, small constant $\epsilon$
\Require clipping bounds $w_{\min}, w_{\max}$
\For{each iteration}
    \State $B \gets \textsc{SampleBatch}(D)$
    \Comment{$B$: batch of prompt-image inputs}

    \For{each input $x \in B$}
        \State $Y_x \gets \textsc{MultiTurnRollout}(\pi_\theta, x, G)$
        \Comment{$Y_x = \{y_{x,1}, \dots, y_{x,G}\}$}
        \State $R_x \gets \textsc{EgoReward}(x, Y_x)$
        \Comment{$R_x = \{r_{x,1}, \dots, r_{x,G}\}$}
        \State $A_x \gets \textsc{GRPOAdv}(R_x)$
        \Comment{$A_x = \{a_{x,1}, \dots, a_{x,G}\}$}
        \State $s_x \gets \textsc{Var}(R_x)$
        \Comment{group statistic}
    \EndFor

    \State $s_{\mathrm{batch}} \gets \textsc{Mean}(\{s_x \mid x \in B\})$

    \For{each input $x \in B$}
        \State $w_x \gets \sqrt{\dfrac{s_x}{s_{\mathrm{g}} + \epsilon}}$
        \State $w_x \gets \textsc{Clip}(w_x, w_{\min}, w_{\max})$
        \State $A_x \gets w_x \cdot A_x$
    \EndFor

    \State $s_{\mathrm{g}} \gets \beta s_{\mathrm{g}} + (1-\beta)s_{\mathrm{batch}}$
    \Comment{EMA update of global statistic}

    \State $\theta \gets \textsc{GRPOUpdate}(\theta, B, \{A_x\}, \pi_{\mathrm{ref}})$
    \Comment{standard GRPO update with KL regularization}
\EndFor
\end{algorithmic}
\end{algorithm}

\subsection{Detailed Breakdown of the Model Pipeline}

\begin{figure*}[t]
    \centering
    \includegraphics[width=\textwidth]{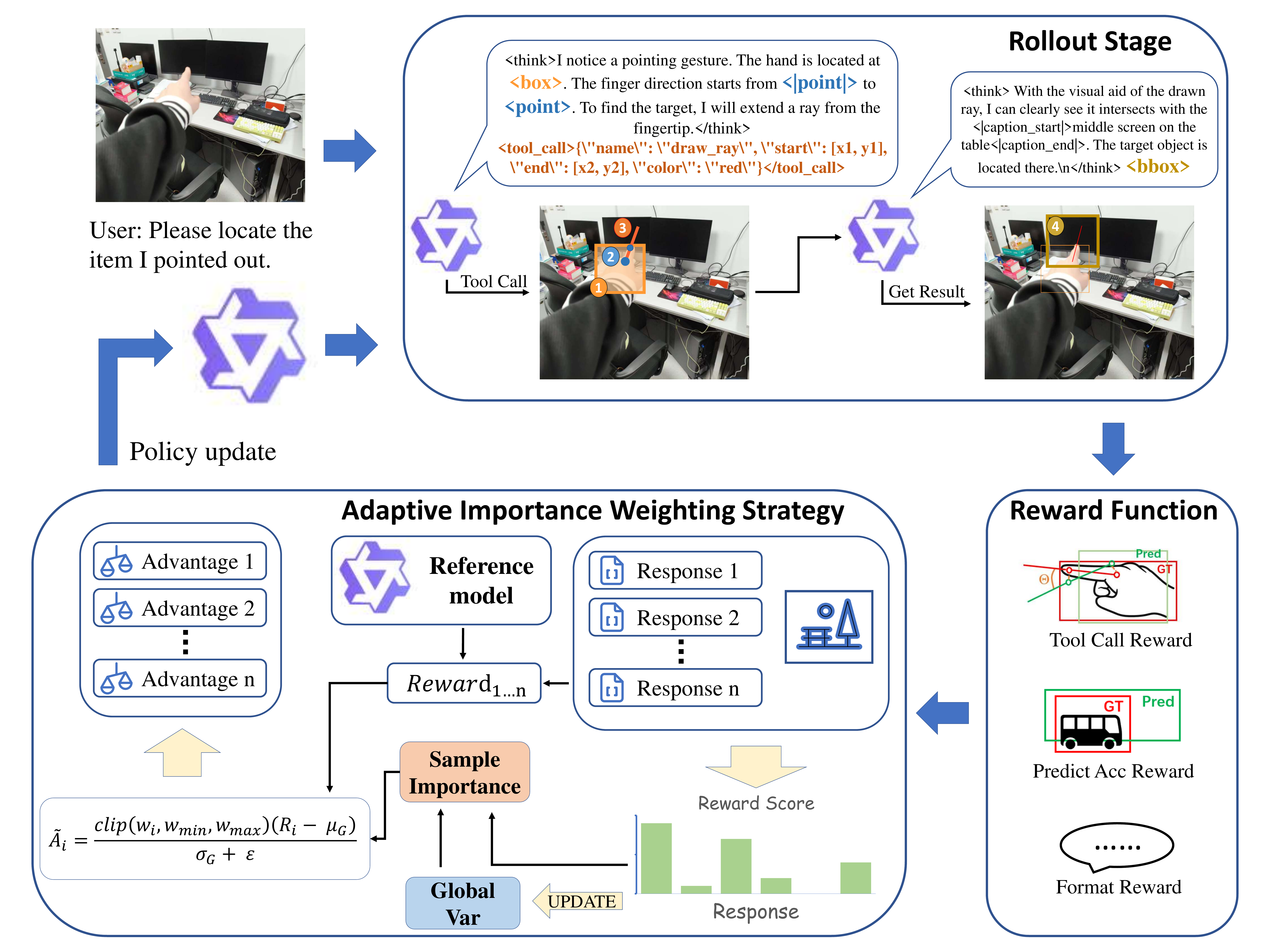}
    \caption{Detailed breakdown of the full model pipeline. Starting from the user query and input image, the model first performs a structured rollout process, including hand localization, pointing keypoint prediction, ray construction via tool calling, and final target grounding. The generated responses are then evaluated by a multi-part reward function, including tool-call reward, prediction accuracy reward, and format reward. Finally, the adaptive importance weighting strategy computes sample-wise importance based on reward statistics, which is used to modulate the policy update during reinforcement learning.}
    \label{fig:pipeline_details}
\end{figure*}

As shown in Figure~\ref{fig:pipeline_details}, our pipeline can be divided into three tightly coupled components. First, in the \emph{rollout stage}, the model receives an egocentric image and a pointing-based query, and generates a structured reasoning trajectory. This trajectory begins with identifying the pointing hand, then estimating the finger direction through keypoint prediction, followed by invoking a drawing tool to explicitly construct a pointing ray. With the visual guidance provided by the ray, the model finally infers the semantic target and outputs the corresponding bounding box.

Second, the generated trajectory is evaluated by a structured \emph{reward function}. Rather than only scoring the final localization result, our design decomposes the reward into multiple components that jointly supervise both intermediate reasoning quality and final prediction correctness. In particular, the tool-call reward measures whether the predicted intermediate geometric reasoning steps are physically meaningful, the prediction accuracy reward evaluates the final grounding result against the ground truth, and the format reward encourages valid and well-structured outputs. This design ensures that the model is optimized not only for correctness but also for reasoning faithfulness.

Third, Figure~\ref{fig:pipeline_details} also illustrates our \emph{adaptive importance weighting strategy}, which is applied during policy optimization. After collecting multiple rollout responses and their corresponding reward scores, we estimate the informativeness of each sampled response according to group-level reward statistics. A global variance term is maintained and used to derive sample importance weights, which further modulate the normalized advantages before the policy update. In this way, more informative rollout groups contribute more strongly to training, while less informative or noisy groups are down-weighted, resulting in a more stable and efficient optimization process.

Overall, Figure~\ref{fig:pipeline_details} provides an intuitive overview of how our method unifies structured visual reasoning, reward-guided learning, and adaptive policy optimization into a single coherent framework.


\subsection{Additional Experimental Analysis}

To provide a more comprehensive understanding of the proposed method, we further present additional experimental analysis from two perspectives: the training dynamics during GRPO optimization and the qualitative comparison against representative baseline MLLMs.

\subsubsection{Training Dynamics during GRPO.}
\begin{figure*}[t]
    \centering
    \includegraphics[width=\textwidth]{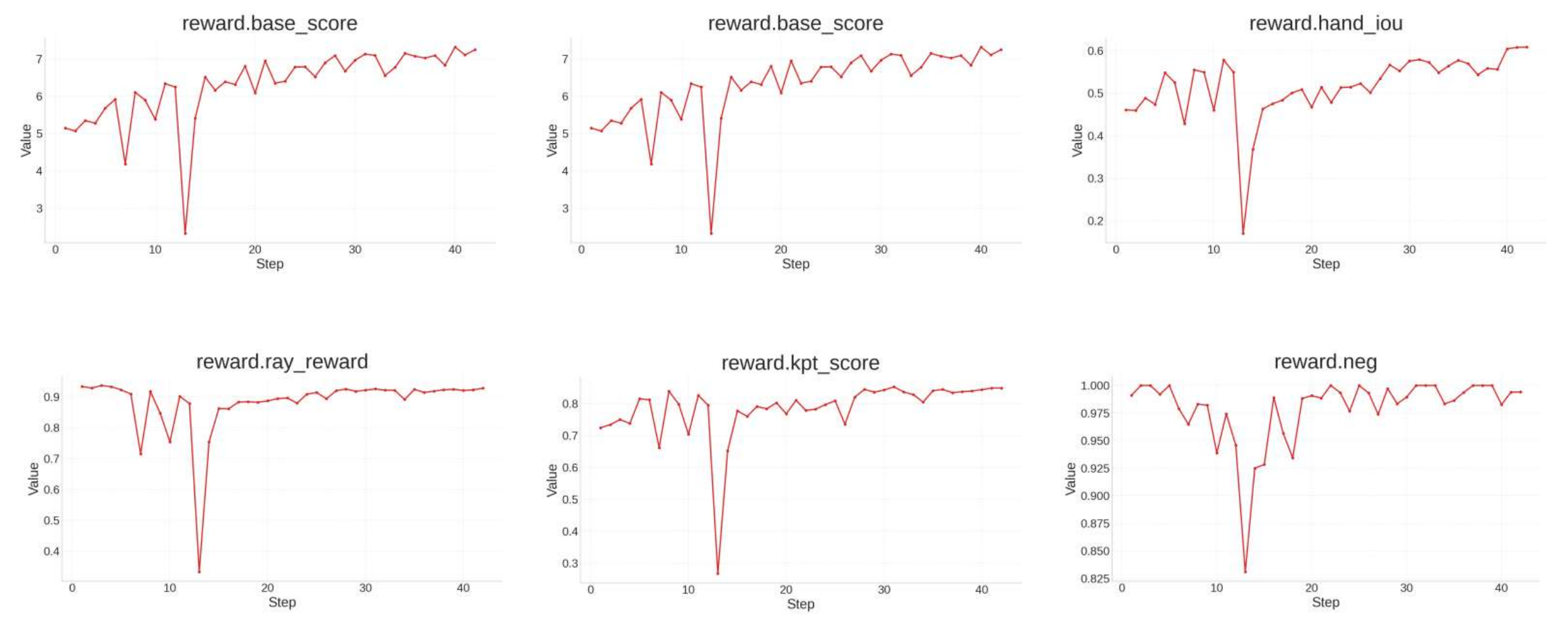}
    \caption{Reward evolution during GRPO training. We visualize the trajectories of the overall base reward and several key reward components, including object IoU reward, hand IoU reward, ray reward, keypoint reward, and negative-sample reward.}
    \label{fig:reward_curve}
\end{figure*}

We first visualize the evolution of the major reward components during reinforcement learning in Figure~\ref{fig:reward_curve}.

As shown in Figure~\ref{fig:reward_curve}, the reward signals exhibit an overall upward trend throughout training, indicating that the policy gradually learns to generate more accurate and more structured reasoning trajectories. In particular, the overall base reward increases steadily as optimization proceeds, suggesting that the model improves jointly on intermediate reasoning quality and final grounding accuracy.

From the perspective of individual reward components, the geometric supervision terms also become progressively more stable. The object IoU reward shows a clear upward tendency, indicating that the final target localization becomes increasingly accurate during GRPO optimization. Similarly, the hand IoU reward improves over training, showing that the model learns to identify the pointing hand more reliably, which provides a stronger spatial anchor for subsequent reasoning. The keypoint reward and ray reward further demonstrate that the model gradually acquires more precise pointing-direction estimation and ray construction ability, which is well aligned with the design motivation of our structured V-CoT framework.

Although several reward terms exhibit transient fluctuations in the early and middle stages of training, all major reward components recover and continue to improve afterward, eventually reaching relatively stable plateaus. This behavior is expected in rollout-based policy optimization and also empirically supports the effectiveness of our adaptive weighting strategy in maintaining robust training under heterogeneous reward quality.

\subsubsection{Qualitative Comparison with Baseline Models.}
\begin{figure*}[!htbp]
    \centering
    \includegraphics[width=\textwidth]{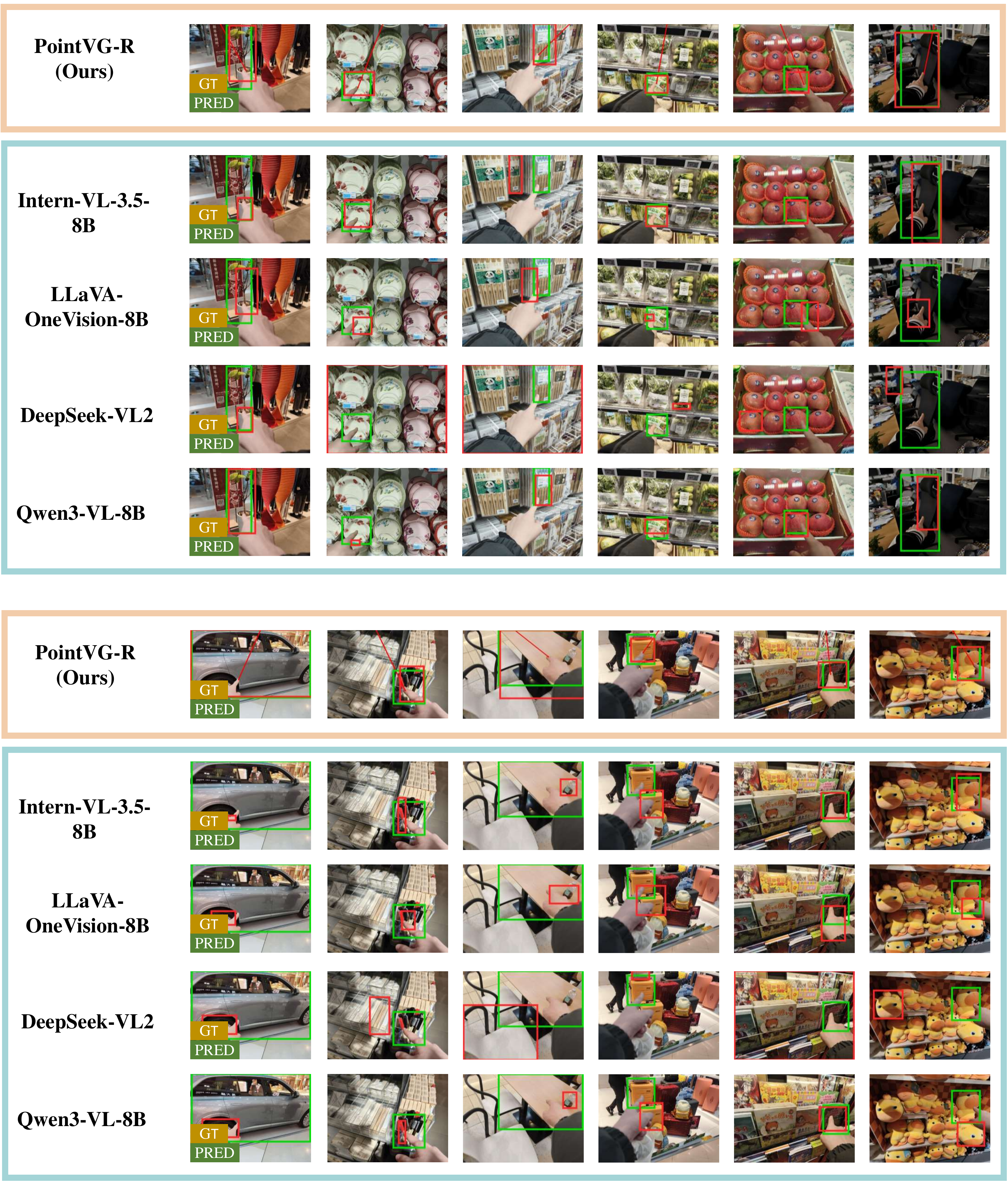}
    \caption{Qualitative comparison between PointVG-R and representative baseline MLLMs. Green boxes denote ground-truth targets and red boxes denote model predictions. Compared with baseline models, PointVG-R produces predictions that are more spatially consistent with the pointing gesture and more accurately aligned with the intended target object.}
    \label{fig:qualitative_comparison}
\end{figure*}

We further provide a qualitative comparison between PointVG-R and several representative baseline MLLMs in Figure~\ref{fig:qualitative_comparison}.
As shown in Figure~\ref{fig:qualitative_comparison}, PointVG-R achieves substantially more accurate and stable localization results than competing models across a wide range of challenging scenarios. 
In most cases, the predicted bounding boxes of PointVG-R closely align with the ground-truth targets, whereas baseline models frequently exhibit systematic failure patterns, including spatial drift, distraction by visually salient objects, and confusion among semantically similar instances. 
These issues become particularly pronounced in cluttered environments, small-object cases, and visually ambiguous scenes, where precise interpretation of pointing cues is essential.

A key advantage of PointVG-R lies in its ability to explicitly model the geometric relationship between the pointing hand and candidate objects. 
Instead of relying on appearance-driven saliency or coarse semantic matching, PointVG-R consistently follows the pointing direction and constrains predictions along the hand-object spatial trajectory, leading to more accurate target disambiguation. 
In contrast, baseline MLLMs tend to attend to nearby salient regions or default to dominant semantic categories, often failing to localize the object that lies along the true pointing direction.

These qualitative findings are consistent with the quantitative results reported in the main paper and further supported by the reward-evolution analysis in Figure~\ref{fig:reward_curve}. 
The comparison suggests that the proposed structured V-CoT reasoning, together with adaptive GRPO optimization, effectively enhances the model’s geometric reasoning capability and reduces reliance on spurious visual priors.

Overall, the additional analyses provide strong evidence for the effectiveness of PointVG-R from both optimization and prediction perspectives. 
The reward curves demonstrate stable and progressive reinforcement learning dynamics under the proposed reward design, while the qualitative results confirm that such improvements translate into more robust and geometry-aware visual grounding behavior, particularly in challenging real-world pointing scenarios.

\subsubsection{Model Comparison}
\begin{figure*}[!h]
    \centering
    \includegraphics[width=\linewidth]{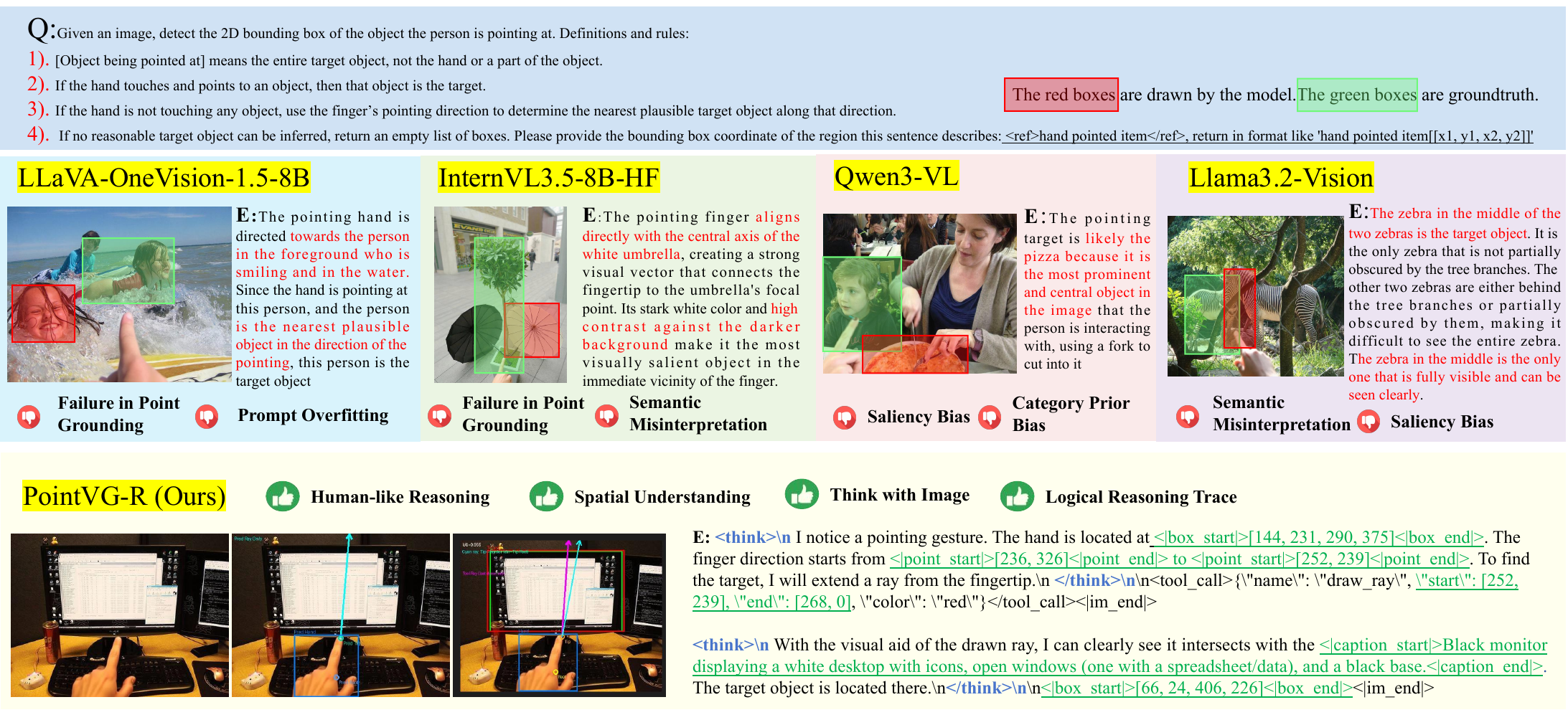}
    \caption{Comparison of different multimodal models on egocentric pointing-based grounding. Existing models (top row) suffer from common failure modes, including point grounding failure, semantic misinterpretation, and saliency bias, often selecting incorrect targets despite clear pointing cues. In contrast, our proposed PointVG-R (bottom row) leverages human-like reasoning and spatial understanding to explicitly model fingertip location and pointing direction, producing interpretable reasoning traces and more accurate grounding results. Red boxes denote model predictions, and green boxes indicate ground truth.}
    \label{fig:supple-comprision}
\end{figure*}

As illustrated in Fig.~\ref{fig:supple-comprision}, existing multimodal large models exhibit several common failure modes in egocentric pointing-based grounding. Specifically, models such as LLaVA-OneVision, InternVL, Qwen3-VL, and Llama3.2-Vision often suffer from \textbf{point grounding failure}, \textbf{semantic misinterpretation}, and \textbf{saliency bias}. For example, some models incorrectly select visually salient regions instead of following the pointing direction, while others fail to correctly align the finger direction with the intended target.

In contrast, our proposed PointVG-R explicitly incorporates \textbf{human-like reasoning} and \textbf{spatial-aware inference}. By modeling the pointing direction and leveraging intermediate reasoning steps, our method can accurately trace the geometric relationship between the fingertip and candidate objects. As shown in the bottom row, PointVG-R produces a clear and interpretable reasoning process, including fingertip localization, directional estimation, and target verification, leading to more precise and robust grounding results.
\section{Dataset Details}
\subsection{Dataset Statistics}

Our dataset consists of egocentric images capturing scenarios where a human hand points to target objects. Each image is annotated with rich multimodal information, including the bounding boxes of the target object and the hand, object category labels, and fingertip keypoints.

\subsubsection{Dataset Overview}
The dataset contains a total of \textbf{15,455} annotated images, split into \textbf{10,837} for training, \textbf{1,546} for validation, and \textbf{3,072} for testing. Among them, \textbf{13,274} images contain both valid hand and target object annotations, which are used for spatial and geometric analysis in the following statistics.

\subsubsection{Object Category Distribution}
Our dataset exhibits strong semantic diversity, covering \textbf{345 distinct object categories} with a total of \textbf{136,468 object instances}. This large number of instances, together with the category richness, reflects the complexity and diversity of real-world egocentric interaction scenarios.

\begin{table}[t]
\centering
\caption{Object category statistics.}
\begin{tabular}{lcc}
\toprule
Metric & Value \\
\midrule
Number of categories & \textbf{345} \\
Number of instances & \textbf{136,468} \\
\bottomrule
\end{tabular}
\end{table}

\subsubsection{Object Density per Image}
To evaluate scene complexity, we analyze the number of objects per image. As shown in Table~\ref{tab:obj_per_image}, each image contains on average \textbf{8.83} objects, with a maximum of \textbf{76} objects. The relatively large standard deviation (\textbf{12.14}) indicates significant variation in scene clutter, suggesting the presence of both simple and highly crowded scenes.

\begin{table}[t]
\centering
\caption{Object count per image statistics.}
\label{tab:obj_per_image}
\begin{tabular}{lccccc}
\toprule
Mean & Median & Min & Max & Std \\
\midrule
\textbf{8.83} & 5 & 1 & \textbf{76} & \textbf{12.14} \\
\bottomrule
\end{tabular}
\end{table}

\subsubsection{Relative Scale of Hands and Objects}
We further analyze the relative scale of hands and target objects, defined as the ratio between bounding box area and image area. 

As shown in Table~\ref{tab:area_stats}, hands occupy a moderate portion of the image (mean \textbf{0.1136}), while target objects tend to be larger (mean \textbf{0.1977}). Notably, a substantial proportion of instances are large (area $>0.1$), with \textbf{44.75\%} for hands and \textbf{61.94\%} for objects, indicating that most interactions occur at close range. In contrast, very small instances are relatively rare (below 3\%), which reduces extreme scale ambiguity.

\begin{table}[t]
\centering
\caption{Relative area statistics of hands and target objects.}
\label{tab:area_stats}
\begin{tabular}{lcccc}
\toprule
 & Mean & Median & Small (\textless 0.01) & Large (\textgreater 0.10) \\
\midrule
Hand & \textbf{0.1136} & 0.0922 & 0.29\% & \textbf{44.75\%} \\
Object & \textbf{0.1977} & 0.1382 & 2.97\% & \textbf{61.94\%} \\
\bottomrule
\end{tabular}
\end{table}

\subsubsection{Spatial Relationship Between Hand and Object}
We analyze the spatial relationship between the hand and the target object by measuring the distance between their bounding box centers.

As shown in Table~\ref{tab:distance_stats}, the average pixel distance is \textbf{374.36}, with a median of 236.90. More importantly, the normalized distance (divided by the image diagonal) has a mean of \textbf{0.2031}, indicating that the hand and target object are typically located in close spatial proximity. This observation is consistent with natural pointing behavior in egocentric interactions.

\begin{table}[t]
\centering
\caption{Hand-object distance statistics.}
\label{tab:distance_stats}
\begin{tabular}{lccc}
\toprule
Metric & Mean & Median & Std \\
\midrule
Pixel distance & \textbf{374.36} & 236.90 & 330.52 \\
Normalized distance & \textbf{0.2031} & 0.1944 & 0.0829 \\
\bottomrule
\end{tabular}
\end{table}

Overall, the dataset demonstrates {high semantic diversity}, {significant variation in scene complexity}, and {realistic spatial configurations} between hands and target objects. These characteristics make it a challenging benchmark for egocentric visual grounding, especially for tasks requiring precise spatial reasoning and multimodal understanding.

\subsection{Annotation Detail of the Dataset}

EgoPoint-CoT is a Visual Chain-of-Thought dataset for egocentric pointing-based visual grounding, built upon EgoPoint Ground. 
To the best of our knowledge, this is the first dataset for pointing understanding that introduces structured intermediate geometric annotations, including hand bounding boxes, fingertip coordinates, 2D pointing rays, and aligned multi-step reasoning descriptions.
Together, these annotations form a complete Visual CoT representation, enabling models to jointly reason about spatial geometry and semantic intent.

To construct the dataset, we first apply MMPose to all images for hand detection and keypoint estimation, obtaining an initial pool of candidate samples. 
Two expert annotators then independently verify the hand bounding boxes and fingertip coordinates, while a third senior annotator resolves disagreements. 
This process results in high annotation reliability, with an inter-annotator agreement of Fleiss' $\kappa = 0.91$.

Based on the refined hand and fingertip annotations, we further construct 2D pointing rays and leverage an LLM to generate step-aligned reasoning descriptions. 
These reasoning chains are subsequently manually reviewed to ensure semantic consistency and correctness, thereby forming a structured reasoning sequence: \textit{hand $\rightarrow$ direction $\rightarrow$ ray $\rightarrow$ target}. 
For negative samples, we additionally design dedicated CoT trajectories (e.g., ``No hand is visible; therefore, no valid target exists.'') to explicitly model invalid pointing scenarios.








\newpage
\bibliographystyle{ACM-Reference-Format}
\bibliography{sample-base}

\end{document}